\documentclass{article}

\usepackage[main, final]{neurips_2026}

\usepackage[utf8]{inputenc}
\usepackage[T1]{fontenc}
\usepackage{hyperref}
\usepackage{url}
\usepackage{booktabs}
\usepackage{amsfonts}
\usepackage{nicefrac}
\usepackage{microtype}
\usepackage{xcolor}
\usepackage{amsmath,amssymb,amsthm,bm,mathtools}
\usepackage{graphicx}
\usepackage{subcaption}
\usepackage{multirow}
\usepackage{enumitem}
\usepackage{algorithm}
\usepackage{algpseudocode}
\usepackage{siunitx}
\usepackage{pdflscape}
\usepackage{rotating}

\usepackage{hyperref}
\hypersetup{
    colorlinks,
    linkcolor={red!50!black},
    citecolor={green!50!black},
    urlcolor={blue!80!black}
}

\newcommand{\best}[1]{\textcolor{teal}{\begingroup\bfseries\boldmath #1\endgroup}}
\providecommand{\second}[1]{\textcolor{purple}{\begingroup\bfseries\boldmath #1\endgroup}}

\usepackage{tikz}
\usetikzlibrary{fit,arrows.meta}
\usetikzlibrary{positioning,calc,arrows.meta}

\title{Flow-Transformed Implicit Processes for Function-Space Variational Inference}

\author{%
  Luis A. Ortega \\
  Aalborg University \\
  \texttt{laoa@cs.aau.dk}\\
  \And
  Andrés R. Masegosa \\
  Aalborg University\\
  \texttt{arma@cs.aau.dk}\\
  \And
  Thomas D. Nielsen \\
  Aalborg University\\
  \texttt{tdn@cs.aau.dk}
}

\begin{document}

\maketitle

\begin{abstract}
Implicit-process priors define distributions over functions through flexible generative mechanisms, making them attractive for Bayesian function-space modelling. However, performing posterior inference with such priors is challenging because their induced function-space distributions are typically not available in closed form. One practical strategy is to approximate the prior using a finite collection of sampled functions, and then represent posterior functions as learned combinations of these samples. Existing approaches commonly place a Gaussian variational distribution over the combination weights. While tractable, this choice limits the shapes of posterior uncertainty that can be represented, especially when the true posterior is asymmetric, heavy-tailed, or multimodal. We propose \emph{Flow-Transformed Implicit Processes} (\textsc{FTIP}), a variational inference method that makes this finite-dimensional function-space approximation more expressive. Instead of using a Gaussian distribution over the combination weights, \textsc{FTIP} uses a normalizing flow to define a richer variational distribution. This induces a flexible posterior distribution over functions while preserving tractable optimization. We train the model using a Black-Box $\alpha$ objective, allowing us to compare mass-covering and mode-seeking variational behaviour. Experiments show that \textsc{FTIP} captures asymmetric and multimodal posterior structure in function space that Gaussian coefficient approximations tend to smooth or collapse.
\end{abstract}

\section{Introduction}

Bayesian neural networks are commonly formulated by placing a prior over weights and approximating the posterior in parameter space \citep{mackay1992practical,neal1996bayesian,blundell2015weight,hernandezlobato2015probabilistic,gal2016dropout}. While computationally convenient, weight-space inference inherits the geometry of neural parameterizations, including symmetries, non-identifiability, strong posterior dependencies, and many-to-one mappings from weights to predictors \citep{sun2019functional,burt2020understanding,wild2022generalized,rudner2022tractable}. Since predictions depend only on the induced input--output map, it is natural to formulate Bayesian inference directly in function space \citep{sun2019functional,ma2019variational,ortega2023deep}.

Function-space inference targets the posterior distribution over functions rather than over parameters. This perspective better aligns inference with prediction, but it also introduces a central computational difficulty: for flexible priors such as Bayesian neural networks and implicit stochastic processes, the induced function-space prior is rarely available through a tractable density \citep{ma2019variational,ortega2026uncertainty}. Practical methods therefore rely on approximations to represent the prior, parameterize the posterior, and optimize a variational objective \citep{sun2019functional,burt2020understanding,wild2022generalized,rudner2022tractable}.

\begin{figure}[t!]
  \centering
  \includegraphics[width=\linewidth]{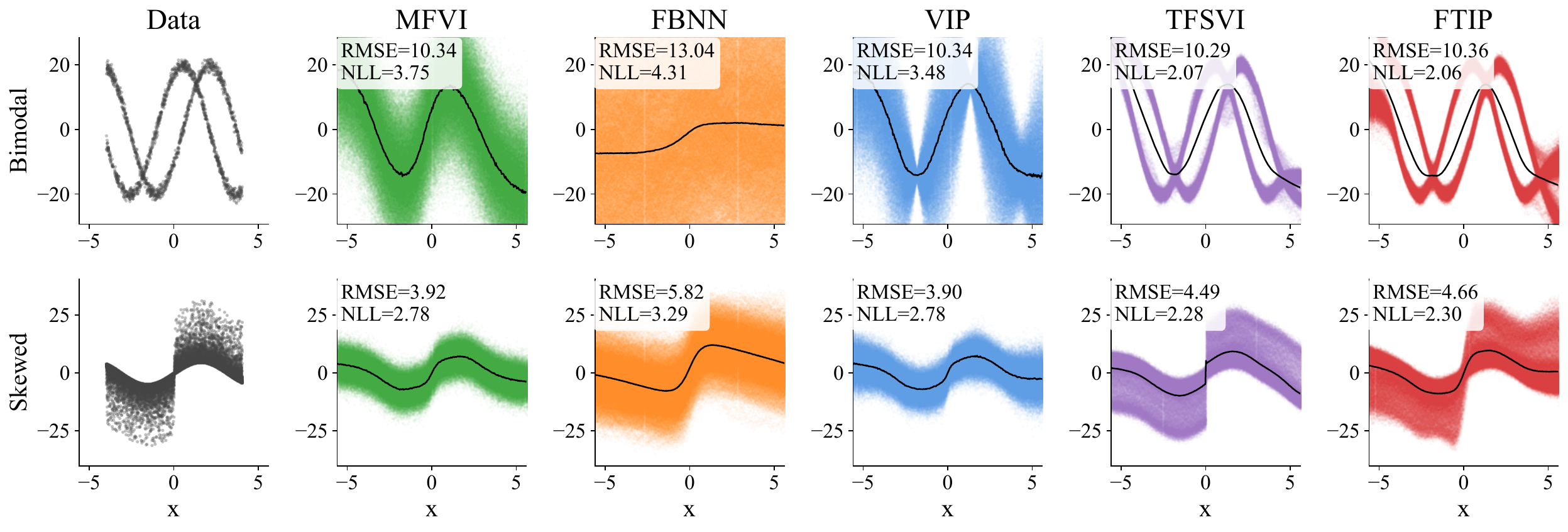}
  \caption{%
    Predictive distributions on two one-dimensional synthetic regression diagnostics. The bimodal dataset, shown in the top row, contains two equally likely conditional branches. The skewed dataset, shown in the bottom row, contains asymmetric log-normal noise whose tail direction changes at $x=0$. Each model panel overlays $1\, 500$ posterior predictive samples on a dense input grid, with the predictive mean shown in black. Test-set \textsc{rmse} and \textsc{nll} are reported in the upper-left corner.
  }
  \label{fig:synthetic-predictive}
\end{figure}

These approximations directly affect both uncertainty representation and predictive performance. Figure~\ref{fig:synthetic-predictive} illustrates several failure modes on simple one-dimensional diagnostics. Mean-field variational inference (\textsc{MFVI}) performs inference in weight space and produces broad, unimodal predictive clouds that do not resolve a bimodal or a skewed structure. Functional Bayesian neural networks (\textsc{FBNN}) \citep{sun2019functional} move the objective into function space, but can still suffer from optimization and approximation difficulties, leading to unstable or poorly calibrated predictions. Variational Implicit Processes (\textsc{VIPs}) \citep{ma2019variational} uses a finite prior-sample surrogate, but its Gaussian posterior over surrogate variables limits expressiveness and tends to smooth over distinct functional explanations. Tractable Function-Space Variational Inference (\textsc{TFSVI}) \citep{rudner2022tractable} improves the functional approximation, but depends on computationally expensive calculations of the neural network Jacobians. These examples motivate a posterior family that remains function-space, sample-forward, and tractable, while being flexible enough to represent non-Gaussian geometry.

We propose \emph{Flow-Transformed Implicit Processes} (\textsc{FTIP}), a function-space variational inference method that combines sample-forward scalability with a more expressive posterior approximation. \textsc{FTIP} starts from a finite collection of prior function samples, preserving the ability to work with implicit priors through forward simulation alone. It then places a normalizing-flow variational distribution over the posterior \citep{rezende2015variational,dinh2017density,papamakarios2021normalizing}. Posterior functions are obtained by sampling base noise, transforming it through an invertible flow, and mapping the transformed variables through the sampled function surrogate.

This construction separates two sources of modelling capacity. The implicit process defines a flexible prior over functions, while the normalizing flow increases the expressiveness of the variational posterior used for inference. As shown in Figure~\ref{fig:synthetic-predictive}, this allows \textsc{FTIP} to recover separated predictive branches in the bimodal setting and asymmetric uncertainty in the skewed setting, while retaining competitive predictive metrics. We train the method using either the standard evidence lower bound or a Black-Box $\alpha$ objective \citep{hernandezlobato2016black}, allowing us to study how posterior expressiveness interacts with mass-covering and mode-seeking variational behaviour.

Our contributions are:
\begin{itemize}[leftmargin=1.2em]
    \item We introduce \textsc{FTIP}, a targeted, nested extension of \textsc{VIP} that augments its finite prior-sample surrogate with a normalizing-flow posterior over surrogate variables.
    \item We show that flow-transformed surrogate posteriors induce richer distributions over functions than Gaussian surrogate posteriors, capturing asymmetric and multimodal posterior structure.
    \item We argue that posterior expressiveness is a central design choice in function-space inference, since approximations to the prior and posterior directly affect uncertainty representation and predictive performance.
    \item We evaluate \textsc{FTIP} on synthetic diagnostics, small to medium regression, and large scale regression benchmarks.
\end{itemize}

\section{Background: Function-Space Variational Inference for Implicit Processes}

Consider data $\mathcal D=\{(\mathbf x_n,y_n)\}_{n=1}^N$ and a Bayesian neural network
with parameters $\bm\theta\in\bm \Theta$. Each parameter value induces a predictor
$f_{\bm\theta}\in\mathcal F$, and the likelihood depends on $\bm\theta$ only
through $p(y| f_{\bm\theta}(\mathbf{x}))$. Standard variational inference places a prior $p(\bm\theta)$ over parameters,
chooses a tractable approximation $q(\bm\theta)$, and maximizes
\begin{equation}
    \mathcal L_{\mathrm{param}}(q(\bm{\theta}))
    =
    \mathbb E_{q(\bm\theta)}[\log p(y_{1:N}\mid \bm\theta)]
    -
    \mathrm{KL}\big(q(\bm\theta)\,\|\,p(\bm\theta)\big).
\end{equation}
While this objective is often tractable, it depends on the chosen
parameterization, even though many parameter values may induce the same
predictive function.

\paragraph{Implicit Processes.}
An implicit process is a stochastic process specified through a sampling
procedure rather than through an explicit density over functions. Let
$\bm{\theta}\sim p(\bm{\theta})$ be a random variable and let
\begin{equation}
    f(\cdot):=g(\cdot,\bm{\theta})
\end{equation}
be a stochastic function generator. For each draw of $\bm{\theta}$, the generator produces
a function $f\in\mathcal F$. Equivalently, defining the measurable map \(T:\mathcal \bm{\Theta}\to\mathcal F, \,T(\bm{\theta}):=g(\cdot,\bm{\theta})\), the process induces a prior distribution over functions by pushforward, $P(f) = T_{\#}p(\bm{\theta})$. Thus, samples from the process are obtained by drawing $\bm{\theta}\sim p(\bm{\theta})$ and
evaluating $g(\cdot,\bm{\theta})$, while the density of $P(f)$ or its marginals is typically unavailable in closed form. Similarly, a variational distribution $q_{\bm\theta}$ induces $ Q(f)=T_{\#}q_{\bm\theta}$. Thus, sampling weights and evaluating the corresponding network is equivalent to
sampling from an implicit prior over functions.

Function-space variational inference instead approximates the posterior directly
as a distribution over functions:
\begin{equation}
    \mathcal L_{\mathrm{func}}(Q(f))
    =
    \mathbb E_{Q(f)}[\log p(y_{1:N}\mid f)]
    -
    \mathrm{KL}\big(Q(f)\,\|\,P(f)\big).
\end{equation}
If $Q(f)=T_{\#}q(\bm{\theta})$ and $P(f)=T_{\#}p(\bm{\theta})$, then the likelihood terms agree and the
KL contracts under the measurable map $T$ \citep{sun2019functional}:
\begin{equation}
    \mathbb E_{q(\bm{\theta})}[\log p(y_{1:N}\mid \bm{\theta})]
    =
    \mathbb E_{Q(f)}[\log p(y_{1:N}\mid f)],
    \qquad
    \mathrm{KL}(Q(f)\,\|\,P(f))
    \leq
    \mathrm{KL}(q(\bm{\theta})\,\|\,p(\bm{\theta})).
\end{equation}
Consequently,
\begin{equation}
    \mathcal L_{\mathrm{func}}(Q(f))
    \geq
    \mathcal L_{\mathrm{param}}(q(\bm{\theta})).
\end{equation}
Function-space inference therefore removes dependence on arbitrary latent parameterizations and produces a tighter ELBO. Its main
obstacle is tractability: for implicit processes, $P(f)$ is sample-defined and
$\mathrm{KL}(Q(f)\,\|\,P(f))$ is generally unavailable in closed form.

\paragraph{Variational Implicit Processes.}
Variational Implicit Processes (\textsc{VIP}) make function-space inference tractable by replacing the implicit prior with a finite-dimensional surrogate constructed from prior samples \citep{ma2019variational}. Draw
\begin{equation}
    f_s(\cdot)=g(\cdot,\bm{\theta}_s),
    \qquad
    \bm{\theta}_s\overset{\mathrm{i.i.d.}}{\sim}p(\bm{\theta}),
    \qquad
    s=1,\dots,S.
\end{equation}
For an input $\mathbf x \in \mathcal{X}$, define
\begin{equation}
    m(\mathbf x)=\frac{1}{S}\sum_{s=1}^S f_s(\mathbf x),
    \qquad
    \phi_s(\mathbf x)=\frac{f_s(\mathbf x)-m(\mathbf x)}{\sqrt{S-1}},
    \qquad
    s=1,\dots,S.
\end{equation}
The \textsc{VIP} surrogate predictor is parametrized by a \textit{coefficient} vector $\bm a\in\mathbb R^S$ as
\begin{equation}
    F(\mathbf x; \bm a)
    =
    m(\mathbf x)
    +
    \sum_{s=1}^S \phi_s(\mathbf x)a_s,
    \qquad
    \bm a=(a_1,\dots,a_S)^\top\in\mathbb R^S,
    \qquad
    p(\bm a)=\mathcal N(\bm 0,\bm I_S).
\end{equation}
The coefficient prior $p(\bm a)$ is distinct from the implicit BNN weight prior used to generate the sampled functions; the latter is specified in Appendix~\ref{app:bnn-prior}. This defines a finite-rank Gaussian process whose empirical mean and covariance match those of the sampled implicit prior on the span of the sampled features.

Posterior inference is performed in coefficient space with a Gaussian variational posterior,
\begin{equation}
    q(\bm a)=\mathcal N(\bm \mu, \bm \Sigma),
\end{equation}
leading to the objective
\begin{equation}
    \mathcal L_{\mathrm{VIP}}
    =
    \mathbb E_{q(\bm a)}
    \big[
        \log p(y_{1:N}\mid F(\cdot;\bm a))
    \big]
    -
    \mathrm{KL}\big(q(\bm a)\,\|\,p(\bm a)\big).
\end{equation}
Hence, \textsc{VIP} can be interpreted as a tractable surrogate for function-space variational inference: the intractable functional KL is replaced by a closed-form coefficient-space KL. Its main limitation is the Gaussian posterior over $\bm a$, which restricts the induced posterior over functions to a unimodal, elliptically structured variational family.

\section{Flow-Transformed Implicit Processes}
\label{sec:ftip}

\begin{figure}[t]
\centering

\begin{minipage}[c]{0.49\textwidth}
\centering
\scalebox{0.8}{
\begin{tikzpicture}[
  >=Latex,
  x=1.0cm, y=1.0cm,
  latent/.style={circle, draw, minimum size=8mm, inner sep=0pt},
  obs/.style={circle, draw, fill=gray!20, minimum size=8mm, inner sep=0pt},
  det/.style={rectangle, draw, rounded corners=2pt, minimum height=6mm, minimum width=6mm, inner sep=4pt},
  plate/.style={draw, rounded corners=2pt, inner sep=10pt},
  lab/.style={font=\small}
]

\node[latent] (t_eps)   at (3,1.0) {$\bm\varepsilon$};
\node[det] (t_theta)   at (2,1) {$\bm\psi$};
\node[det]    (t_a)     at (2.6,-1.5) {$\bm a=T_{\bm\psi}(\bm\varepsilon)$};

\node[det] (t_x)   at (4.5,0.1) {$\mathbf x_n$};
\node[obs] (t_y)   at (6.8,-1.5) {$y_n$};

\node[latent] (t_z) at (6.8,2) {$\bm \theta_s$};
\node[det]    (t_f) at (6.8,0.1) {$f_{n,s}=g(\mathbf x_n,\bm \theta_s)$};

\draw[->] (t_x) -- (t_f);
\draw[->] (t_z) -- (t_f);
\draw[->] (t_f) -- (t_y);
\draw[->] (t_a) -- (t_y);
\draw[->] (t_eps) -- (t_a);
\draw[->] (t_theta) -- (t_a);

\node[plate, fit=(t_x) (t_f) (t_y)] (t_plateN) {};
\node[plate, fit=(t_z) (t_f)] (t_plateS) {};
\node[lab, anchor=south east] at (t_plateN.south east) {$N$};
\node[lab, anchor=north east] at (t_plateS.north east) {$S$};

\end{tikzpicture}}
\end{minipage}
\hfill
\begin{minipage}[c]{0.5\textwidth}
\centering
\includegraphics[width=\linewidth]{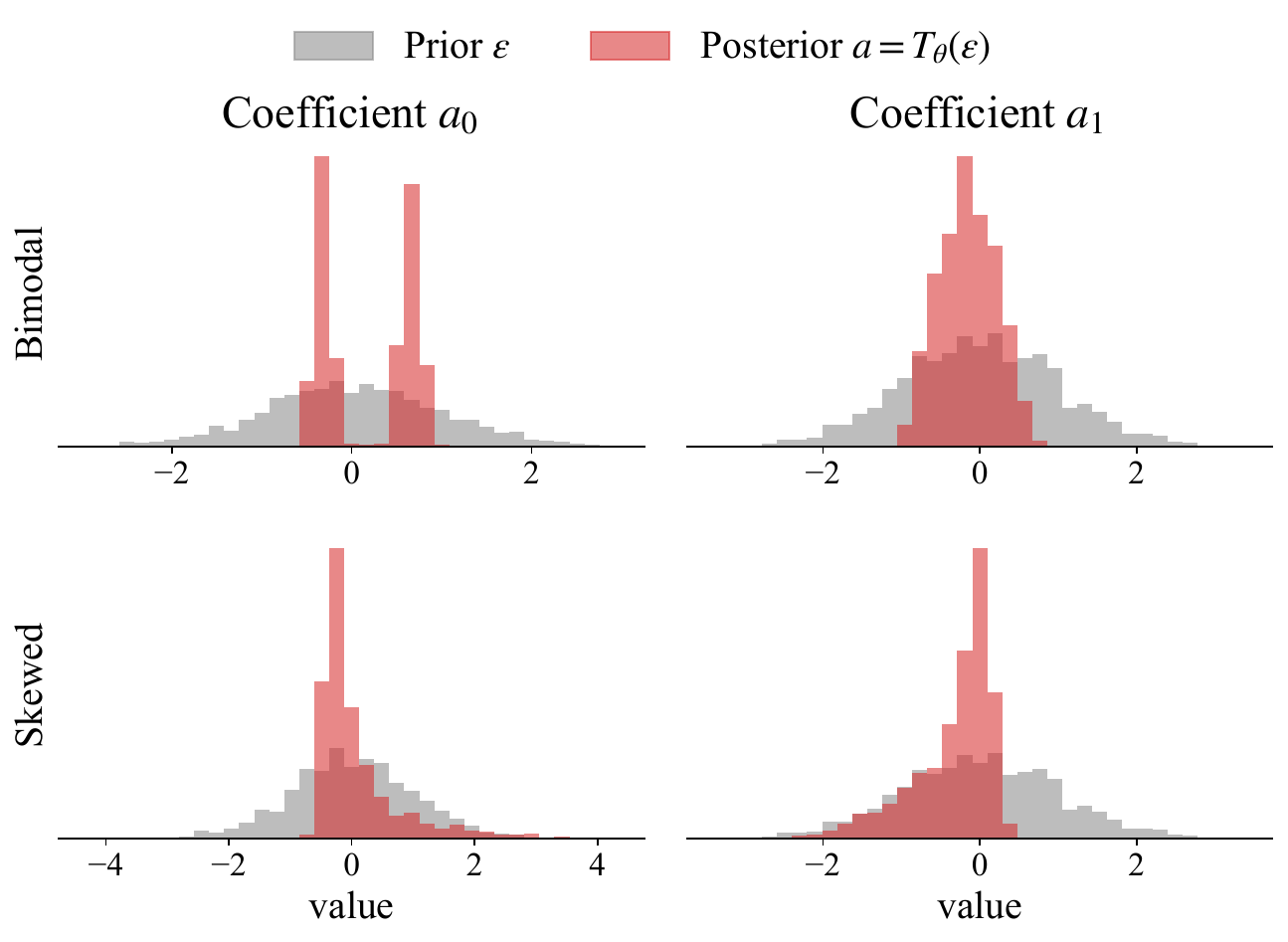}
\end{minipage}

\caption{\small Left: graphical model for FTIP. Global coefficients are generated by a flow, $\bm a=T_{\bm{\psi}}(\bm \varepsilon)$, and combined with sampled implicit-process features. Right: example distribution induced by the flow over the coefficient space (or over a selected projection of the flow output).}
\label{fig:ftip_graphical_model}
\end{figure}
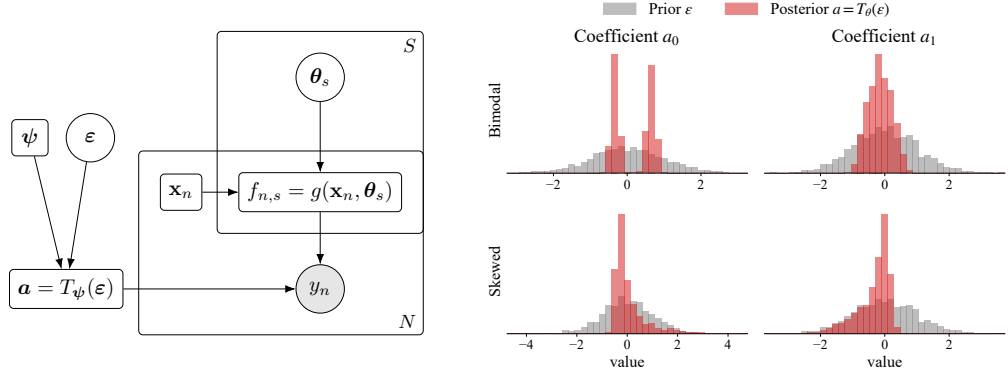

\textsc{FTIP} is explicitly designed to retain the scalability and sample-forward tractability that make \textsc{VIP} attractive for inference with implicit priors, while substantially enlarging the flexibility of the predictive posteriors it can represent, e.g.,  non-Gaussian, asymmetric, and multimodal predictive distributions over functions. Recall that the \textsc{VIP} surrogate represents functions as
\begin{equation}
    F(\mathbf x;\bm a)=m(\mathbf x)+\sum_{s=1}^S \phi_s(\mathbf x)a_s,
    \qquad
    p(\bm a)=\mathcal N(\bm 0,\bm I_S),
\end{equation}
where the coefficients $\bm a\in\mathbb R^S$ parameterize functions in the span of the sampled prior features. Standard \textsc{VIP} places a Gaussian variational posterior over these coefficients. In contrast, \textsc{FTIP} defines the posterior by transforming Gaussian base noise through an invertible map. Specifically, for $\bm \varepsilon\sim\mathcal N(\bm 0,\bm I_S)$, we set $\bm a=T_{\bm \psi}(\bm \varepsilon)$, where $T_{\bm \psi}:\mathbb R^S\to\mathbb R^S$ is a normalizing flow \citep{rezende2015variational,dinh2017density,papamakarios2021normalizing}. The induced density is tractable through the change-of-variables formula
\begin{equation}
    \log q_{\bm \psi}(\bm a)
    =
    \log p(\bm \varepsilon)
    -
    \log\left|
        \det
        \frac{\partial T_{\bm \psi}(\bm \varepsilon)}
             {\partial \bm \varepsilon}
    \right|,
    \qquad
    \bm a=T_{\bm \psi}(\bm \varepsilon).
\end{equation}

The resulting variational objective is the same coefficient-space ELBO used by \textsc{VIP}, but with the Gaussian posterior replaced by the flow posterior:
\begin{equation}
    \mathcal L_{\mathrm{FTIP}}
    =
    \mathbb E_{q_{\bm \psi}(\bm a)}
    \big[
        \log p(y_{1:N}\mid F(\cdot;\bm a))
    \big]
    -
    \mathrm{KL}\big(q_{\bm \psi}(\bm a)\,\|\,p(\bm a)\big).
    \label{eq:ftip-elbo}
\end{equation}
Using the reparameterization $\bm a=T_{\bm \psi}(\bm \varepsilon)$, this objective can be optimized by Monte Carlo samples from the base distribution:
\begin{equation}
    \mathcal L_{\mathrm{FTIP}}
    =
    \mathbb E_{\bm \varepsilon\sim\mathcal N(\bm 0,\bm I_S)}
    \left[
        \log p\big(y_{1:N}\mid F(\cdot;T_{\bm \psi}(\bm \varepsilon))\big)
        +
        \log p\big(T_{\bm \psi}(\bm \varepsilon)\big)
        -
        \log q_{\bm \psi}\big(T_{\bm \psi}(\bm \varepsilon)\big)
    \right].
\end{equation}
Thus, \textsc{FTIP} retains the tractable surrogate prior and coefficient-space KL structure of \textsc{VIP}, while replacing the elliptically contoured Gaussian posterior with a more flexible pushforward distribution.

Posterior prediction is obtained by sampling coefficients from the flow posterior and evaluating the induced surrogate functions. For regression, samples $\bm a^{(k)}\sim q_{\bm \psi}(\bm a)$ define
\begin{equation}
    F^{(k)}(\mathbf x_\star)
    =
    m(\mathbf x_\star)+\sum_{s=1}^S \phi_s(\mathbf x_\star)a_s^{(k)},
\end{equation}
and the predictive distribution is approximated as the Monte Carlo mixture
\begin{equation}
    p(y_\star\mid \mathbf x_\star,\mathcal D)
    \approx
    \frac{1}{K}\sum_{k=1}^K
    p\big(y_\star\mid F^{(k)}(\mathbf x_\star)\big).
\end{equation}
This makes the predictive distribution a consequence of the variational posterior over functions: richer coefficient distributions induce richer distributions over surrogate functions. Figure~\ref{fig:ftip_graphical_model} shows the graphical model of FTIPs and the empirical distribution of two of the coefficients in the Bimodal and Skewed data of Figure~\ref{fig:synthetic-predictive}. 

In our experiments, the flow $T_{\bm\psi}$ is implemented as an initial affine transformation followed by rational-quadratic spline coupling layers \citep{durkan2019neural} interleaved with invertible $1\times1$ LU mixing layers \citep{kingma2018glow}:
\begin{equation}
    T_{\bm \psi}
    =
    T_{\mathrm{LU}}^{(L)}
    \circ
    T_{\mathrm{RQS}}^{(L)}
    \circ\cdots\circ
    T_{\mathrm{LU}}^{(1)}
    \circ
    T_{\mathrm{RQS}}^{(1)}
    \circ
    T_{\mathrm{aff}} .
\end{equation}
Here $T_{\mathrm{aff}}(\bm \varepsilon)=\bm M\bm \varepsilon+\bm b$ provides a trainable Gaussian base posterior when the nonlinear coupling layers are disabled. It induces $q(\bm a)=\mathcal N(\bm b,\bm M\bm M^\top)$, exactly the full-covariance Gaussian coefficient posterior of \textsc{VIP}. Thus, \textsc{FTIP} is a nested extension of \textsc{VIP}: the only change to the posterior family is the addition of nonlinear invertible transformations after the affine map. The spline coupling layers then provide flexible monotone transformations of subsets of the coefficient dimensions, while the LU layers mix dimensions between coupling blocks. Together, these transformations preserve exact sampling, inversion, and density evaluation, while allowing $q_{\bm \psi}(\bm a)$ to represent non-Gaussian geometry such as skewness, heavy tails, and separated modes. Full details of the affine map, spline parameterization, LU log-determinants, initialization, and density computation are given in Appendix~\ref{app:flow-details}.

\subsection{Training objective}\label{sec:training}

As an alternative to the ELBO in Equation~\eqref{eq:ftip-elbo}, we train \textsc{FTIP} using a Black-Box $\alpha$ objective \citep{hernandezlobato2016black}. This objective is shared by \textsc{VIP} and \textsc{FTIP}, and we match $\alpha$ in every comparison. As $\alpha\to0$, its likelihood term converges to the expected log-likelihood of the standard Monte Carlo ELBO; at $\alpha=1$, it becomes the log-likelihood of a posterior-predictive Monte Carlo mixture, allowing different function samples to explain different plausible outcomes. Importantly, the objective does not create multimodality by itself: \textsc{VIP} uses the same objective, but its Gaussian coefficient posterior may be too restrictive to exploit multiple functional explanations. This objective modifies the likelihood aggregation across posterior samples while retaining the same coefficient-space KL regularization. For a minibatch $\mathcal B$, posterior samples $\bm a^{(k)}\sim q_{\bm \psi}(\bm a)$, and shorthand $F_n^{(k)}=F(\mathbf x_n;\bm a^{(k)})$, we optimize
\begin{equation}
\mathcal L_{\alpha}
=
\frac{N_{\mathrm{data}}}{N_{\mathrm{batch}}}
\sum_{n\in\mathcal B}
\frac{1}{\alpha}
\log
\left[
\frac{1}{K}
\sum_{k=1}^{K}
\exp\left(
    \alpha \log p(y_n\mid F_n^{(k)})
\right)
\right]
-
\mathrm{KL}\big(q_{\bm \psi}(\bm a)\,\|\,p(\bm a)\big)
-
R_{\mathrm{prior}}(\bm\theta).
\label{eq:bbalpha}
\end{equation}
The coefficient-space KL regularizes the posterior coefficients. The separate $R_{\mathrm{prior}}(\bm\theta)$ is the evidence-style prior-adaptation safeguard inherited from \textsc{VIP}; it regularizes the parameters of the implicit prior and is not an \textsc{FTIP} posterior regularizer in weight space. Full derivations are given in Appendix~\ref{app:objective-details}.

In every matched comparison, $\alpha$ and $S$ are shared by \textsc{VIP} and \textsc{FTIP}; $\alpha$ is selected on validation data using the application-relevant metric, and $S$ is treated as a validation-selected basis-capacity parameter. We use a matched two-branch continuation protocol and report a controlled IID-versus-antithetic comparison in Appendices~\ref{app:warmstart} and~\ref{app:antithetic-sampling}.

\section{Related Work}

Function-space variational inference was developed to avoid some of the geometric difficulties of weight-space Bayesian neural networks, such as parameter symmetries, non-identifiability, and strong posterior dependencies that do not necessarily correspond to predictive uncertainty \citep{sun2019functional,burt2020understanding,wild2022generalized,rudner2022tractable}. 

\paragraph{Functional Bayesian Neural Networks (fBBNs)} Early functional Bayesian neural network methods, such as \textsc{fBNNs}, formulate variational inference directly over the stochastic process induced by a Bayesian neural network, replacing a weight-space KL with an approximate functional KL evaluated on finite measurement sets \citep{sun2019functional}. This gives a conceptually clean function-space objective, but its quality depends on the chosen measurement sets and on finite-dimensional approximations to an infinite-dimensional process. 

\paragraph{Variational Implicit Processes (VIPs)} VIPs make inference tractable by constructing a finite-dimensional surrogate from sampled prior functions \citep{ma2019variational}. Posterior inference is performed over global surrogate coefficients, whose pushforward defines an approximate posterior over functions. This sample-forward construction is scalable and broadly applicable to implicit priors, but standard \textsc{VIP} restricts the coefficient posterior to be Gaussian, limiting the geometry of the induced function-space posterior. 

\paragraph{Tractable Function-Space Variational Inference (TFSVI)} More recently, TFSVI made functional objectives more practical for
Bayesian neural networks by approximating 
using a local linearization of the network. In particular, they use a
first-order Taylor expansion around the mean parameters and compute the induced
Gaussian function distribution from network Jacobians evaluated on finite context
sets \citep{rudner2022tractable}. This avoids the intractable exact
function-space KL, but introduces a Jacobian-based computational bottleneck and
ties the approximation quality to both the local linearization and the choice of
context points. 

These methods strengthen the practical case for function-space inference, but they still depend on finite measurement sets or surrogate representations whose posterior expressiveness and computational cost must be carefully balanced. Our work follows this line of function-space methods, but focuses specifically on increasing posterior expressiveness within a sample-forward surrogate.

\paragraph{Expressive variational posteriors}
Normalizing flows enrich variational families by transforming simple base distributions through invertible maps with tractable Jacobian determinants \citep{rezende2015variational,dinh2017density,papamakarios2021normalizing}. Coupling flows provide efficient sampling and density evaluation, while rational-quadratic spline couplings and invertible $1\times1$ mixing layers improve local flexibility and dimension mixing \citep{durkan2019neural,kingma2018glow}. These methods are commonly used to obtain more expressive approximate posteriors in finite-dimensional latent-variable models. Transformed GPs also use flows to enrich both a GP prior and its variational approximation \citep{maronas2021transforming}. \textsc{FTIP} uses normalizing flows in a different role: the flow is placed over the surrogate variables of a function-space approximation, and its pushforward through the sampled function surrogate induces a richer posterior distribution over functions. Thus, rather than increasing the expressiveness of the implicit prior sampler, \textsc{FTIP} increases the expressiveness of the variational posterior used for inference.

\section{Experiments}

We evaluate predictive performance using root mean square error (RMSE), negative log-likelihood (NLL),
continuous ranked probability score (CRPS), and the centered quantile metric
(CQM)~\citep{ortega_variational_2024}. However, our main focus is on NLL and CQM. The goal of \textsc{FTIP} is
not only to improve point prediction, but to represent richer posterior
predictive distributions. This distinction is important because RMSE depends
only on the predictive mean and can therefore favor models that average over
distinct explanations. As illustrated in Figure~\ref{fig:synthetic-predictive},
such averaging is undesirable in the bimodal diagnostic, where the conditional
distribution has two separated branches, and in the skewed diagnostic, where the
predictive uncertainty is asymmetric. In both cases, a model may obtain a
reasonable RMSE while assigning little probability mass to the observed target
under the full predictive distribution.

NLL and CQM are better aligned with this objective. NLL evaluates the density
assigned to the observed targets and therefore directly rewards predictive
distributions that place probability mass on the appropriate modes and tails.
CQM complements this by measuring the quality of predictive quantiles, making it
sensitive to calibration errors that arise when a posterior is too symmetric,
too diffuse, or collapses onto an averaged explanation. We still report RMSE
and CRPS for completeness, but treat them as secondary metrics in settings where
the conditional distribution is multimodal or skewed. In particular, CRPS can be
closely tied to RMSE in calibrated Gaussian settings, and we discuss this
relationship in Appendix~\ref{app:crps}.

The codebase for the experiments can be found in \url{https://github.com/Ludvins/ImplicitProcessZoo}.

\subsection{Synthetic regression diagnostics}

We first consider two one-dimensional regression diagnostics designed to test whether the posterior predictive can represent non-Gaussian conditional structure. The bimodal dataset contains two equally likely sinusoidal branches, while the skewed dataset contains asymmetric log-normal noise whose tail direction changes at $x=0$. Full dataset definitions and training details are given in Appendix~\ref{app:experimental-details}.

Figure~\ref{fig:synthetic-predictive} shows posterior predictive samples for \textsc{MFVI}, \textsc{FBNN}, \textsc{VIP}, \textsc{TFSVI}, and \textsc{FTIP}. On the bimodal task, \textsc{MFVI}, \textsc{FBNN} and \textsc{VIP} produces Gaussian or effectively unimodal posterior predictive approximations that tend to average across the two branches or inflate uncertainty to cover both. In contrast, \textsc{FTIP} and \textsc{TFSVI} produce predictive posteriors that occupy the separated predictive modes. On the skewed task, \textsc{FTIP} and \textsc{TFSVI} better capture asymmetric predictive uncertainty, while \textsc{MFVI}, \textsc{FBNN} and \textsc{VIP} remain more symmetric and diffuse. Note how the NLL metric properly measures these approximation capacities, in contrast to RMSE. 

These diagnostic results illustrate the main effect of the flow-transformed coefficient posterior: it preserves the finite prior-sample surrogate of \textsc{VIP}, but induces a richer posterior distribution over functions that can represent both multimodality and skewness.  {Appendix~\ref{app:broader-baselines} reports five-seed HMC and deep-ensemble diagnostics and explains the different inferential targets of these methods.}

\subsection{UCI regression}

\begin{table}[t]
    \centering
    \small
    \caption{UCI regression test negative log-likelihood (NLL). Mean results and standard deviation over $5$ seeds is reported. Median training time per iteration is also shown. Best results and methods within the corresponding error range are highlighted.}
    \label{tab:uci_simpler_a05_nll}
    \scalebox{0.9}{
    \begin{tabular}{lccccc}
        \toprule
        \textbf{Dataset} & \textbf{VIP} & \textbf{FTIP} & \textbf{FBNN} & \textbf{MFVI} & \textbf{TFSVI} \\
        \midrule
        Boston &
        \bm{$2.65 \pm 0.07$} &
        \bm{$2.64 \pm 0.14$} &
        $3.46 \pm 0.01$ &
        $3.62 \pm 0.04$ &
        $3.04 \pm 0.14$ \\

        Concrete &
        \bm{$3.48 \pm 0.04$} &
        \bm{$3.48 \pm 0.03$} &
        $4.07 \pm 0.09$ &
        $4.22 \pm 0.06$ &
        $3.91 \pm 0.00$ \\

        Kin8nm &
        \bm{$-0.295 \pm 0.006$} &
        \bm{$-0.282 \pm 0.014$} &
        $0.046 \pm 0.006$ &
        $0.048 \pm 0.005$ &
        $-0.037 \pm 0.064$ \\

        Naval &
        \bm{$-4.479 \pm 0.214$} &
        \bm{$-4.232 \pm 0.237$} &
        $-2.797 \pm 0.006$ &
        $-2.797 \pm 0.006$ &
        $-2.970 \pm 0.097$ \\

        Power &
        \bm{$2.82 \pm 0.02$} &
        \bm{$2.83 \pm 0.04$} &
        $4.03 \pm 0.06$ &
        $4.00 \pm 0.01$ &
        $2.93 \pm 0.05$ \\

        Wine &
        \bm{$0.96 \pm 0.03$} &
        \bm{$0.95 \pm 0.01$} &
        $1.14 \pm 0.04$ &
        $1.14 \pm 0.04$ &
        $1.03 \pm 0.04$ \\

        Yacht &
        \bm{$2.58 \pm 0.19$} &
        \bm{$2.49 \pm 0.19$} &
        $3.83 \pm 0.28$ &
        $4.04 \pm 0.01$ &
        $3.65 \pm 0.01$ \\
        \midrule
        ms/iter & $12.9$ & $17.0$ & $6.4$ & $5.3$ & $49.5$\\
        \bottomrule
    \end{tabular}}
\end{table}
\begin{figure}[t]
    \centering
    
    \begin{minipage}[c]{0.45\linewidth}
        \scalebox{0.7}{
        \begin{tabular}{llcccc}
    \toprule
    Model & $\alpha$ & \multicolumn{4}{c}{Energy} \\
    \cmidrule(lr){3-6}
     &  & RMSE $\downarrow$ & NLL $\downarrow$ & CRPS $\downarrow$ & CQM $\downarrow$ \\
    \midrule
    \multirow{2}{*}{VIP} & $0.5$ & $\bm{2.99 \pm 0.03}$ & $2.11 \pm 0.01$ & $1.42 \pm 0.01$ & $0.035 \pm 0.013$ \\
     & $1.0$ & $3.03 \pm 0.04$ & $2.05 \pm 0.10$ & $\bm{1.38 \pm 0.12}$ & $0.061 \pm 0.009$ \\
    \midrule
    \multirow{2}{*}{FTIP} & $0.5$ & $\bm{3.00 \pm 0.03}$ & $2.03 \pm 0.00$ & $\bm{1.38 \pm 0.01}$ & $0.033 \pm 0.009$ \\
     & $1.0$ & $3.10 \pm 0.30$ & $\bm{1.66 \pm 0.04}$ & $\bm{1.38 \pm 0.14}$ & $\bm{0.026 \pm 0.007}$ \\
    \midrule
    Model & $\alpha$ & \multicolumn{4}{c}{Protein} \\
    \cmidrule(lr){3-6}
     &  & RMSE $\downarrow$ & NLL $\downarrow$ & CRPS $\downarrow$ & CQM $\downarrow$ \\
    \midrule
    \multirow{2}{*}{VIP} & $0.5$ & $\bm{4.96 \pm 0.00}$ & $3.02 \pm 0.00$ & $2.83 \pm 0.02$ & $0.020 \pm 0.006$ \\
     & $1.0$ & $5.09 \pm 0.06$ & $2.98 \pm 0.02$ & $\bm{2.79 \pm 0.03}$ & $0.042 \pm 0.005$ \\
    \midrule
    \multirow{2}{*}{FTIP} & $0.5$ & $\bm{4.96 \pm 0.01}$ & $3.02 \pm 0.00$ & $\bm{2.82 \pm 0.01}$ & $0.020 \pm 0.006$ \\
     & $1.0$ & $5.45 \pm 0.02$ & $\bm{2.60 \pm 0.00}$ & $2.90 \pm 0.02$ & $\bm{0.009 \pm 0.001}$ \\
    \bottomrule
\end{tabular}
}
    \end{minipage}\hfill
    \begin{minipage}[c]{0.4\linewidth}
        \includegraphics[width=0.8\linewidth]{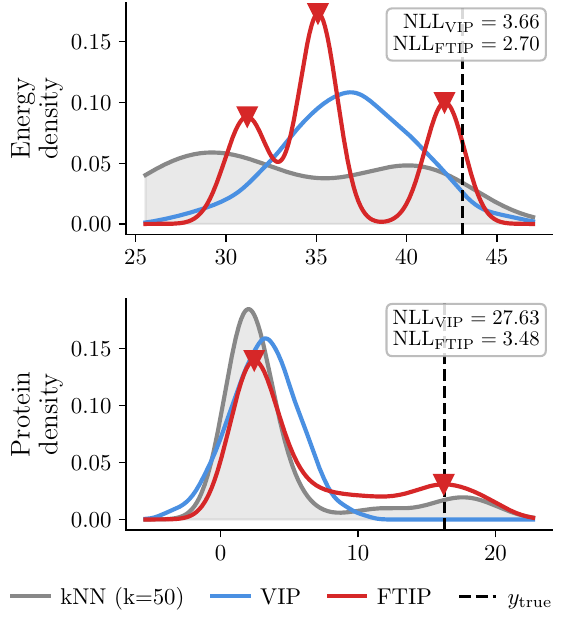}
    \end{minipage}
    \caption{
    Energy and Protein results for \textsc{VIP} and \textsc{FTIP} with $\alpha\in\{0.5,1.0\}$. 
    Left: test RMSE, NLL, CRPS, and CQM. Best results are highlighted.
    Right: predictive densities at selected test inputs, together with a $k$-NN density estimate and the observed target value.
    }
\label{fig:uci_multimodal_alpha}
\end{figure}

We evaluate \textsc{FTIP} on standard UCI regression benchmarks \citep{Dua2019uci} against
\textsc{VIP}, \textsc{FBNN}, \textsc{MFVI}, and \textsc{TFSVI}. We report RMSE,
NLL, CRPS, and CQM over five seeds, for $\alpha\in\{0.5,1.0\}$. Full results
are given in Table~\ref{tab:uci_simpler_full} (Appendix); Table~\ref{tab:uci_simpler_a05_nll}
shows representative NLL results for $\alpha=0.5$.

Table~\ref{tab:uci_simpler_a05_nll} shows that \textsc{FTIP} improves over
\textsc{FBNN}, \textsc{MFVI}, and \textsc{TFSVI} across all reported datasets,
while remaining comparable to \textsc{VIP}. In several cases the two methods are
within the same error range. We hypothesize that these datasets might have conditional
predictive distributions that are close to unimodal and approximately Gaussian,
so the Gaussian coefficient posterior used by \textsc{VIP} is already
sufficient. In that regime, \textsc{FTIP} mainly reproduces the same predictive
distribution as VIP through flow samples rather than exploiting strongly non-Gaussian
posterior structure.

Energy and Protein show a different behaviour. As shown in
Figure~\ref{fig:uci_multimodal_alpha}, when $\alpha=1.0$, \textsc{FTIP}
substantially improves NLL and CQM over \textsc{VIP}. This may reflect
multimodality or skewness in the conditional target distributions. The
predictive-density plots support this interpretation by comparing the model
predictive densities with empirical $k$-NN density estimates: \textsc{VIP} produces smoother
Gaussian-like densities, whereas \textsc{FTIP} assigns probability mass more
adaptively near high-density regions. See Appendix~\ref{app:objective-details} for an intuition on how \(\alpha\) affects these results.

\subsection{Large-scale Regression}

We next evaluate \textsc{FTIP} on the YearPredictionMSD benchmark \citep{BertinMahieux2011msd}, a larger
regression task with substantially more observations than the UCI datasets.
All methods use the same two-hidden-layer Bayesian neural network architecture. Results are reported in
Table~\ref{tab:year_results}. \textsc{FTIP} obtains the best NLL, CRPS, and CQM, while \textsc{VIP} achieves
the lowest RMSE. This pattern is consistent with the preceding experiments:
\textsc{VIP} remains highly competitive as a point predictor, but its Gaussian
coefficient posterior is less effective at representing the full predictive
distribution. By contrast, \textsc{FTIP} slightly sacrifices RMSE while assigning
better-calibrated probability mass to the targets, leading to a substantial
improvement in NLL and CQM.

The result is particularly important because it shows that the benefit of the
flow-transformed posterior is not limited to small synthetic or UCI datasets.
On YearPredictionMSD, \textsc{FTIP} retains the sample-forward scalability of
\textsc{VIP} while improving distributional prediction quality. Compared with
\textsc{TFSVI}, which relies on Jacobian-based function-space approximations,
\textsc{FTIP} achieves better uncertainty metrics without requiring neural
network Jacobians.

\begin{table}[t]
    \centering
    \small
    \caption{%
    Test performance on large-scale regression Year dataset. {\color{teal} \textbf{Best}} and {\color{purple}\textbf{second-to-best}} are high-lighted. All methods report mean\,$\pm$\,std over 5 seeds.
    }
    \label{tab:year_results}
    \scalebox{0.8}{
    \begin{tabular}{lcccc}
        \toprule
        \textbf{Method} & \textbf{RMSE} $\downarrow$ & \textbf{NLL} $\downarrow$ & \textbf{CRPS} $\downarrow$ & \textbf{CQM} $\downarrow$ \\
        \midrule
        VIP
          & \best{$10.23 \pm 0.06$}
          & \second{$3.655 \pm 0.015$}
          & \second{$5.326 \pm 0.046$}
          & \second{$0.029 \pm 0.003$}\\
        FTIP
          & \second{$10.31 \pm 0.05$}
          & \best{$3.407 \pm 0.007$}
          & \best{$5.209 \pm 0.028$}
          & \best{$0.010 \pm 0.006$}\\
        FBNN
          & $12.11 \pm 1.52$
          & $3.888 \pm 0.175$
          & $6.863 \pm 1.054$
          & $0.075 \pm 0.043$\\
        MFVI
          & $10.80 \pm 0.00$
          & $3.798 \pm 0.000$
          & $5.778 \pm 0.003$
          & $0.046 \pm 0.000$\\
        TFSVI
          & $10.90 \pm 0.12$
          & $3.720 \pm 0.110$
          & $5.741 \pm 0.085$
          & $0.042 \pm 0.004$\\
        \bottomrule
    \end{tabular}}
\end{table}

\subsection{Pedestrian future path-length prediction}
\label{sec:pedestrian-path-length}

\begin{figure}[t]
    \centering
    \begin{minipage}[c]{0.49\linewidth}
        \centering
        \includegraphics[width=\linewidth]{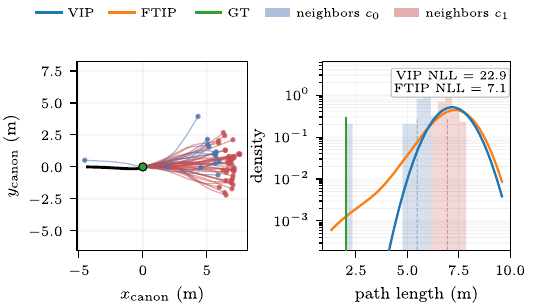}
        \label{fig:placeholder}
    \end{minipage}\hfill
    \begin{minipage}[c]{0.49\linewidth}
        \centering
        \scalebox{0.8}{
        \begin{tabular}{lcc}
    \toprule
    Method & RMSE $\downarrow$ & CRPS $\downarrow$ \\
    \midrule
    VIP   & \bm{$0.996 \pm 0.04$} & \bm{$0.519\pm 0.03$} \\
    FTIP  & \bm{$1.026\pm 0.05$} & \bm{$0.531\pm 0.02$} \\
    MFVI  & $2.192\pm 0.03$      & $1.278\pm 0.02$      \\
    FBNN  & $2.221\pm 0.03$      & $1.300\pm 0.03$      \\
    \midrule
    Method & NLL $\downarrow$ & CQM $\downarrow$ \\
    \midrule
    VIP   & $1.42\pm 0.09$       & \bm{$0.024\pm 0.009$} \\
    FTIP  & \bm{$1.26\pm 0.06$}  & \bm{$0.036\pm 0.008$} \\
    MFVI  & $2.20\pm 0.01$       & $0.066\pm 0.012$      \\
    FBNN  & $2.30\pm 0.05$       & $0.058\pm 0.021$      \\
    \bottomrule
\end{tabular}

}
        
    \end{minipage}
    \caption{%
    Pedestrian path-length prediction. Left: a representative test prefix with
    nearest-neighbor future trajectories from two clusters and the corresponding
    path-length predictive densities from \textsc{VIP} and \textsc{FTIP}; right:
    test performance across methods.
    }
    \label{fig:pedestrian_pathlen}
\end{figure}

We evaluate \textsc{FTIP} on a scalar trajectory-forecasting task derived from
ETH/UCY pedestrian trajectories \citep{Pellegrini2009eth,Lerner2007ucy}. Given
the last $T_{\mathrm{obs}}=8$ observed positions $p_t\in\mathbb R^2$, the target
is the total length of the next $T_{\mathrm{pred}}=12$ steps using an LSTM network. Since the same
observed prefix can lead to different future walking speeds, $p(y|\mathbf x)$
is often skewed or multimodal. Figure~\ref{fig:pedestrian_pathlen} illustrates
this with a representative input and ground truth (GT) future, clustered neighbor (near input) futures, and the induced
path-length predictive densities.

The results in Figure~\ref{fig:pedestrian_pathlen} show that \textsc{VIP} and
\textsc{FTIP}, which infer in low-dimensional surrogate coefficient space,
achieve roughly $2\times$ lower RMSE and CRPS than \textsc{MFVI} and
\textsc{FBNN}, whose diagonal Gaussian posterior is placed over the full
parameter vector. \textsc{TFSVI} is omitted due to the heavy Jacobian overhead
on the LSTM backbone. Among surrogate methods, \textsc{FTIP} obtains the best
NLL, indicating sharper and better-calibrated predictive distributions. This
supports the role of the flow posterior: when path-length uncertainty is
not Gaussian-like, added posterior expressiveness improves uncertainty quality beyond a
Gaussian surrogate posterior.

\paragraph{Prior misspecification diagnostic}
We additionally study whether the flow posterior can compensate for a
misspecified Gaussian-process base prior. This experiment, reported in
Appendix~\ref{app:prior-misspecification}, uses a bimodal regression task where
the exact GP posterior remains Gaussian and therefore averages over the two
predictive branches. The results show that \textsc{FTIP} can induce a
non-Gaussian posterior over functions from the same finite prior-sample
surrogate, allowing it to represent separated functional explanations that the
Gaussian GP posterior cannot capture.

\paragraph{Classification}
We also report binary and multiclass classification experiments in
Appendix~\ref{app:classification}. These tasks are interpreted as tests of
calibration and posterior approximation quality rather than as continuous
multimodality diagnostics, since Bernoulli and categorical likelihoods already
define non-Gaussian predictive distributions over labels. In these experiments,
\textsc{FTIP} improves over \textsc{VIP} mainly in NLL, ECE, and Brier score.

\section{Discussion and Future Work}

We introduced \textsc{FTIP}, a  {targeted, nested} flow-transformed extension of variational
implicit processes that increases posterior expressivity while preserving the
finite function-space surrogate construction of \textsc{VIP}. Standard
\textsc{VIP} places a Gaussian posterior over coefficients associated with
sampled prior functions, yielding a tractable and scalable approximation but
restricting the geometry of the induced posterior over functions. \textsc{FTIP}
replaces this Gaussian coefficient posterior with a normalizing-flow posterior,
allowing the surrogate to represent non-Gaussian uncertainty while keeping
inference in the same low-dimensional coefficient space.

The experiments suggest that this additional flexibility is most useful when the
continuous predictive distribution is not well described by a unimodal or
symmetric approximation. On synthetic bimodal and skewed regression diagnostics,
\textsc{FTIP} captures separated predictive branches and asymmetric uncertainty
patterns that are smoothed out by Gaussian variational baselines. On UCI
regression, large-scale regression, and pedestrian path-length prediction,
\textsc{FTIP} remains competitive in point-prediction metrics while improving
likelihood-based and calibration-sensitive metrics in settings where richer
posterior structure is beneficial. Classification results in
Appendix~\ref{app:classification} further suggest that the flow posterior can
improve calibration metrics beyond regression, although these
tasks do not test continuous output-space multimodality in the same way as the
regression diagnostics.

Overall, \textsc{FTIP} highlights posterior expressivity as an important design
choice in function-space variational inference. Rather than modifying the
implicit prior sampler, it enriches the posterior family placed over the
surrogate coefficients. This can partially compensate for restrictive Gaussian
surrogate posteriors and, in some cases, for misspecified base priors, while
retaining the sampling-based flexibility that makes implicit-process methods
attractive.

\paragraph{Limitations } The additional expressiveness of \textsc{FTIP} has a real computational cost in parameters, memory, training time, and sampled prediction; detailed profiling is reported in Appendix~\ref{app:computational-profiling}. \textsc{FTIP} is not expected to dominate \textsc{VIP} uniformly: we recommend \textsc{VIP} when validation indicates that its Gaussian coefficient posterior is adequate, and \textsc{FTIP} only when improvements in distribution-sensitive metrics justify the extra cost. The experiments do not establish scalability to contemporary foundation-scale architectures or high-dimensional structured-output distributions; in particular, the pedestrian experiment predicts scalar path length rather than a full trajectory distribution. The method also remains limited by its finite prior-sample basis. Finally, the synthetic HMC and ensemble diagnostics do not imply universal superiority over MCMC, SGLD, SWAG, or alternative ensemble objectives.

\textbf{Future Work} Lighter flow architectures, adaptive criteria
for activating nonlinear flow layers, and regularization schemes could improve
stability and reduce computational overhead. Another promising direction is to
develop diagnostics that determine when a Gaussian surrogate posterior is
sufficient and when a flow-transformed posterior is necessary. Combining
flow-transformed posteriors with more adaptive prior-sample bases may further
reduce the gap between the finite surrogate and the original implicit process.

\subsection*{Acknowledgments} This work was supported by grants (VIL50093, VIL78584) from VILLUM FONDEN and by a grant (NNF23400085294) from the Novo Nordisk Foundation.

\bibliographystyle{plainnat}
\bibliography{references}

\newpage 
\appendix

\section{Societal Impact}

This work is methodological and focuses on improving uncertainty estimation in
Bayesian function-space inference. Better calibrated predictive distributions
may benefit forecasting and scientific modelling, especially when multiple
outcomes are plausible. However, \textsc{FTIP} does not address data bias,
privacy, fairness, or distribution shift by itself. In applications involving
people, such as trajectory prediction, deployment would require appropriate
consent, privacy protection, calibration checks, and domain-specific validation.
The added computational cost should also be justified by a clear need for
non-Gaussian predictive uncertainty.

\section{Details of the \textsc{FTIP} Flow}
\label{app:flow-details}

This appendix describes the normalizing-flow posterior used in
Section~\ref{sec:ftip}. Throughout, the base variable is
\[
    \bm\varepsilon \sim p_0(\bm\varepsilon)
    =
    \mathcal N(\bm 0,\bm I_S),
\]
and the coefficient vector is generated as
\[
    \bm a = T_{\bm\psi}(\bm\varepsilon),
    \qquad
    \bm a\in\mathbb R^S .
\]
The induced coefficient posterior is denoted by
\[
    q_{\bm\psi}(\bm a)
    =
    (T_{\bm\psi})_{\#}p_0(\bm\varepsilon),
\]
and is used in the surrogate predictor
\[
    F(\mathbf x;\bm a)
    =
    m(\mathbf x)
    +
    \sum_{s=1}^{S}\phi_s(\mathbf x)a_s .
\]

The flow used in the experiments has the form
\begin{equation}
    T_{\bm\psi}
    =
    T_{\mathrm{LU}}^{(L)}
    \circ
    T_{\mathrm{RQS}}^{(L)}
    \circ
    \cdots
    \circ
    T_{\mathrm{LU}}^{(1)}
    \circ
    T_{\mathrm{RQS}}^{(1)}
    \circ
    T_{\mathrm{aff}} .
    \label{eq:appendix-flow-composition}
\end{equation}
The trainable parameters of all affine, coupling, and mixing layers are collected
in $\bm\psi$.

\paragraph{Initial affine map.}
The first transformation is a trainable affine map
\begin{equation}
    \bm h^{(0)}
    =
    T_{\mathrm{aff}}(\bm\varepsilon)
    =
    \bm M\bm\varepsilon+\bm b,
    \label{eq:appendix-affine-map}
\end{equation}
where $\bm M\in\mathbb R^{S\times S}$ is nonsingular and
$\bm b\in\mathbb R^S$. When the subsequent nonlinear layers are initialized
close to the identity, this affine map gives a Gaussian coefficient posterior,
\[
    \bm a \approx \bm M\bm\varepsilon+\bm b,
    \qquad
    q_{\bm\psi}(\bm a)
    \approx
    \mathcal N(\bm b,\bm M\bm M^\top).
\]
The log-determinant contribution of the affine map is
\begin{equation}
    \log
    \left|
    \det
    \frac{\partial T_{\mathrm{aff}}(\bm\varepsilon)}
         {\partial \bm\varepsilon}
    \right|
    =
    \log |\det \bm M|.
\end{equation}

\paragraph{Rational-quadratic spline coupling layers.}
Each rational-quadratic spline coupling layer partitions its input vector
$\bm h\in\mathbb R^S$ into two groups of coordinates,
\[
    \bm h =
    (\bm h_{\mathcal I_\ell},\bm h_{\mathcal J_\ell}),
\]
and transforms only the coordinates in $\mathcal J_\ell$:
\begin{equation}
    \widetilde{\bm h}_{\mathcal I_\ell}
    =
    \bm h_{\mathcal I_\ell},
    \qquad
    \widetilde h_j
    =
    \tau_{\ell,j}
    \big(
        h_j;
        \eta_{\ell,j}(\bm h_{\mathcal I_\ell})
    \big),
    \qquad
    j\in\mathcal J_\ell .
    \label{eq:appendix-rqs-coupling}
\end{equation}
Here $\eta_{\ell,j}$ is a neural conditioner and $\tau_{\ell,j}$ is an
elementwise monotone rational-quadratic spline \citep{durkan2019neural}.

For each transformed coordinate $j\in\mathcal J_\ell$, the conditioner outputs
unnormalized bin widths, bin heights, and interior derivatives,
\begin{equation}
    \eta_{\ell,j}(\bm h_{\mathcal I_\ell})
    =
    \left(
        \omega_{\ell,j,1:R},
        \nu_{\ell,j,1:R},
        \rho_{\ell,j,1:R-1}
    \right).
\end{equation}
These are mapped to valid spline parameters by
\begin{align}
    w_{\ell,j,r}
    &=
    w_{\min}
    +
    (2B_{\mathrm{tail}}-Rw_{\min})
    \frac{\exp(\omega_{\ell,j,r})}
         {\sum_{m=1}^{R}\exp(\omega_{\ell,j,m})},
    \\
    v_{\ell,j,r}
    &=
    v_{\min}
    +
    (2B_{\mathrm{tail}}-Rv_{\min})
    \frac{\exp(\nu_{\ell,j,r})}
         {\sum_{m=1}^{R}\exp(\nu_{\ell,j,m})},
    \\
    d_{\ell,j,r}
    &=
    d_{\min}
    +
    \operatorname{softplus}(\rho_{\ell,j,r}),
    \qquad r=1,\dots,R-1,
\end{align}
where $w_{\min},v_{\min},d_{\min}>0$. The boundary derivatives are fixed as
\[
    d_{\ell,j,0}
    =
    d_{\ell,j,R}
    =
    1 .
\]
The input and output knots are
\begin{align}
    \kappa_{\ell,j,0}
    &=
    -B_{\mathrm{tail}},
    &
    \kappa_{\ell,j,r}
    &=
    -B_{\mathrm{tail}}
    +
    \sum_{m=1}^{r}w_{\ell,j,m},
    \\
    \lambda_{\ell,j,0}
    &=
    -B_{\mathrm{tail}},
    &
    \lambda_{\ell,j,r}
    &=
    -B_{\mathrm{tail}}
    +
    \sum_{m=1}^{r}v_{\ell,j,m}.
\end{align}
Hence
\[
    \kappa_{\ell,j,R}
    =
    \lambda_{\ell,j,R}
    =
    B_{\mathrm{tail}} .
\]
Outside the interval $[-B_{\mathrm{tail}},B_{\mathrm{tail}}]$, identity tails
are used:
\begin{equation}
    \tau_{\ell,j}(u)=u,
    \qquad
    |u|>B_{\mathrm{tail}} .
\end{equation}

For
\[
    u=h_j
    \in
    [\kappa_{\ell,j,r-1},\kappa_{\ell,j,r}],
\]
define
\begin{equation}
    \xi
    =
    \frac{
        u-\kappa_{\ell,j,r-1}
    }{
        w_{\ell,j,r}
    },
    \qquad
    s_{\ell,j,r}
    =
    \frac{
        v_{\ell,j,r}
    }{
        w_{\ell,j,r}
    } .
\end{equation}
The spline transformation is
\begin{equation}
    \tau_{\ell,j}(u)
    =
    \lambda_{\ell,j,r-1}
    +
    v_{\ell,j,r}
    \frac{
        s_{\ell,j,r}\xi^2
        +
        d_{\ell,j,r-1}\xi(1-\xi)
    }{
        s_{\ell,j,r}
        +
        \big(
            d_{\ell,j,r}
            +
            d_{\ell,j,r-1}
            -
            2s_{\ell,j,r}
        \big)
        \xi(1-\xi)
    } .
    \label{eq:appendix-rqs-transform}
\end{equation}
Because all widths, heights, and derivatives are positive, the transformation is
strictly monotone. Its inverse is computed by locating the corresponding output
bin and solving the resulting scalar quadratic equation.

The Jacobian of a coupling layer is triangular, so its log-determinant is
\begin{equation}
    \log
    \left|
    \det
    \frac{
        \partial \widetilde{\bm h}
    }{
        \partial \bm h
    }
    \right|
    =
    \sum_{j\in\mathcal J_\ell}
    \log
    \left|
    \frac{
        \partial \tau_{\ell,j}(h_j)
    }{
        \partial h_j
    }
    \right| .
\end{equation}
For
$u\in[\kappa_{\ell,j,r-1},\kappa_{\ell,j,r}]$, the derivative is
\begin{equation}
    \frac{
        \partial \tau_{\ell,j}(u)
    }{
        \partial u
    }
    =
    \frac{
        s_{\ell,j,r}^{2}
        \left[
            d_{\ell,j,r}\xi^2
            +
            2s_{\ell,j,r}\xi(1-\xi)
            +
            d_{\ell,j,r-1}(1-\xi)^2
        \right]
    }{
        \left[
            s_{\ell,j,r}
            +
            \big(
                d_{\ell,j,r}
                +
                d_{\ell,j,r-1}
                -
                2s_{\ell,j,r}
            \big)
            \xi(1-\xi)
        \right]^2
    } .
    \label{eq:appendix-rqs-derivative}
\end{equation}

\paragraph{Invertible $1\times1$ LU mixing layers.}
Between spline coupling layers, we apply an invertible linear mixing layer
\begin{equation}
    \widetilde{\bm h}
    =
    \bm W_\ell \bm h+\bm c_\ell ,
    \label{eq:appendix-lu-layer}
\end{equation}
where $\bm c_\ell\in\mathbb R^S$ and $\bm W_\ell$ is nonsingular. We parameterize
\[
    \bm W_\ell
    =
    \bm P_\ell \bm L_\ell \bm U_\ell ,
\]
where $\bm P_\ell$ is a permutation matrix, $\bm L_\ell$ is lower triangular
with unit diagonal, and $\bm U_\ell$ is upper triangular with nonzero diagonal.
The inverse is
\begin{equation}
    \bm h
    =
    \bm W_\ell^{-1}
    \left(
        \widetilde{\bm h}-\bm c_\ell
    \right),
\end{equation}
and the log-determinant is
\begin{equation}
    \log |\det \bm W_\ell|
    =
    \sum_{j=1}^{S}
    \log |(\bm U_\ell)_{jj}| .
    \label{eq:appendix-lu-logdet}
\end{equation}
These layers mix coefficient dimensions between coupling transformations so that
successive coupling layers can affect all coordinates of $\bm a$.

\paragraph{Flow density.}
Let
\[
    \bm h^{(0)}
    =
    T_{\mathrm{aff}}(\bm\varepsilon),
\]
and, for $\ell=1,\dots,L$,
\begin{align}
    \bm h^{(2\ell-1)}
    &=
    T_{\mathrm{RQS}}^{(\ell)}
    \big(
        \bm h^{(2\ell-2)}
    \big),
    \\
    \bm h^{(2\ell)}
    &=
    T_{\mathrm{LU}}^{(\ell)}
    \big(
        \bm h^{(2\ell-1)}
    \big).
\end{align}
The final coefficient vector is
\[
    \bm a
    =
    \bm h^{(2L)}
    =
    T_{\bm\psi}(\bm\varepsilon).
\]
By the change-of-variables formula,
\begin{align}
    \log q_{\bm\psi}(\bm a)
    &=
    \log p_0(\bm\varepsilon)
    -
    \log|\det \bm M|
    \nonumber\\
    &\quad
    -
    \sum_{\ell=1}^{L}
    \log
    \left|
    \det
    \frac{
        \partial T_{\mathrm{RQS}}^{(\ell)}
    }{
        \partial \bm h^{(2\ell-2)}
    }
    \right|
    -
    \sum_{\ell=1}^{L}
    \log|\det \bm W_\ell|,
    \label{eq:appendix-flow-density}
\end{align}
where
\[
    \bm\varepsilon
    =
    T_{\bm\psi}^{-1}(\bm a).
\]
Equivalently, for a reparameterized sample
$\bm a=T_{\bm\psi}(\bm\varepsilon)$, the same expression gives the exact
log-density needed in the variational objective.

The coefficient-space KL appearing in the \textsc{FTIP} objective is
\begin{equation}
    \mathrm{KL}
    \big(
        q_{\bm\psi}(\bm a)
        \,\|\,
        p(\bm a)
    \big)
    =
    \mathbb E_{q_{\bm\psi}(\bm a)}
    \left[
        \log q_{\bm\psi}(\bm a)
        -
        \log p(\bm a)
    \right],
    \qquad
    p(\bm a)
    =
    \mathcal N(\bm 0,\bm I_S).
    \label{eq:appendix-coefficient-kl}
\end{equation}
In practice, this expectation is estimated with Monte Carlo samples
\[
    \bm\varepsilon^{(k)}
    \sim
    \mathcal N(\bm 0,\bm I_S),
    \qquad
    \bm a^{(k)}
    =
    T_{\bm\psi}(\bm\varepsilon^{(k)}).
\]

\paragraph{Initialization.}
The flow is initialized close to a Gaussian coefficient posterior. The affine
map in Eq.~\eqref{eq:appendix-affine-map} provides the initial Gaussian
location and scale. The spline coupling layers are initialized close to the
identity by making the bins approximately uniform and the derivatives close to
one,
\[
    \tau_{\ell,j}(u)
    \approx
    u .
\]
The LU mixing layers are initialized near well-conditioned identity or
permutation-like transformations. Thus, early training behaves similarly to a
Gaussian posterior over the coefficients $\bm a$, while subsequent optimization
can deform $q_{\bm\psi}(\bm a)$ into a more expressive non-Gaussian posterior.
\section{Objective and Prior-Regularization Details}
\label{app:objective-details}

\paragraph{Black-Box $\alpha$ objective.}
The Black-Box $\alpha$ objective is motivated by the $\alpha$-divergence
\begin{equation}
    D_\alpha[p\|q]
    =
    \frac{1}{\alpha(1-\alpha)}
    \left(
        1-\int p(u)^\alpha q(u)^{1-\alpha}\,du
    \right),
    \qquad
    \alpha\neq0,1.
\end{equation}
The limiting cases recover the two KL divergences:
\begin{equation}
    \lim_{\alpha\to0}D_\alpha[p\|q]
    =
    \mathrm{KL}(q\|p),
    \qquad
    \lim_{\alpha\to1}D_\alpha[p\|q]
    =
    \mathrm{KL}(p\|q).
\end{equation}
The likelihood aggregation term in Eq.~\eqref{eq:bbalpha} satisfies
\begin{equation}
    \frac{1}{\alpha}
    \log
    \left[
        \frac{1}{K}
        \sum_{k=1}^{K}
        \exp\left(
            \alpha \log p(y_n\mid F_n^{(k)})
        \right)
    \right]
    \xrightarrow[\alpha\to0]{}
    \frac{1}{K}
    \sum_{k=1}^{K}
    \log p(y_n\mid F_n^{(k)}),
\end{equation}
so Eq.~\eqref{eq:bbalpha} reduces to the Monte Carlo ELBO as $\alpha\to0$.
In this limit, every posterior sample contributes through its log-likelihood,
so samples that explain a different mode of the data can be heavily penalized
on observations from the current mode. This tends to favour a single compromise
solution or a mode-seeking approximation when the posterior predictive
distribution is multimodal.

For $\alpha>0$, the log-mean-exp term increasingly emphasizes posterior samples
that assign high likelihood to each observation. In particular, when
$\alpha=1$, the likelihood term becomes
\begin{equation}
    \log
    \left[
        \frac{1}{K}
        \sum_{k=1}^{K}
        p(y_n\mid F_n^{(k)})
    \right],
\end{equation}
which is the log-likelihood of a Monte Carlo mixture over posterior function
samples. This is better suited to multimodal predictive distributions: different
samples can explain different branches or modes, and an observation is assigned
high probability as long as some sufficiently weighted posterior samples place
mass near it. The $\alpha=1$ objective therefore does not create multimodality
by itself, but it allows an expressive variational family such as the
flow-transformed posterior in \textsc{FTIP} to use multiple functional
explanations instead of collapsing them into a single Gaussian-like prediction.

\paragraph{ Prior adaptation inherited from \textsc{VIP}.}
 {Let $\bm\theta$ denote the parameters of the implicit prior defined in Appendix~\ref{app:bnn-prior}. Following \citet{ma2019variational}, a hierarchical treatment places a hyperprior $p(\bm\theta)$ and a variational posterior $q(\bm\theta)$, yielding the schematic objective}
\begin{equation}
 {
\mathcal L_{\alpha}^{\mathrm{VIP}}
=
\mathcal L_{\alpha,\mathrm{lik}}
-
\mathrm{KL}\big(q(\bm a)\,\|\,p(\bm a)\big)
-
\mathrm{KL}\big(q(\bm\theta)\,\|\,p(\bm\theta)\big).}
\end{equation}
 {The last KL regularizes the learned prior parameters, whereas the coefficient-space KL regularizes the posterior coefficients. \textsc{VIP} uses the empirical-Bayes approximation $q(\bm\theta)=\delta(\bm\theta-\bm\theta_q)$. Its evidence-style safeguard follows from the standard bound}
\begin{equation}
 {
\log q_{\mathrm{GP}}(\bm y\mid\bm X)
\geq
\mathbb E_{q(\bm\theta)}
\left[\log q_{\mathrm{GP}}(\bm y\mid\bm X,\bm\theta)\right]
-
\mathrm{KL}\big(q(\bm\theta)\,\|\,p(\bm\theta)\big),}
\end{equation}
 {where $q_{\mathrm{GP}}(\bm y\mid\bm X,\bm\theta)$ is the marginal likelihood under the finite-sample GP surrogate. Under the Dirac approximation, this motivates}
\begin{equation}
 {
-\mathrm{KL}\big(q(\bm\theta)\,\|\,p(\bm\theta)\big)
\approx
-\log q_{\mathrm{GP}}(\bm y\mid\bm X,\bm\theta)+C,}
\end{equation}
 {and hence $R_{\mathrm{prior}}(\bm\theta_q)\approx\log q_{\mathrm{GP}}(\bm y\mid\bm X,\bm\theta_q)+C$ in Eq.~\eqref{eq:bbalpha}. This safeguard penalizes adapted priors that overfit through the full surrogate covariance, a failure mode not exposed by the diagonal per-observation likelihood terms. It is inherited from \textsc{VIP}, is used in the regression studies, and is set to zero for both methods in classification.}

\section{Training Algorithm}
\label{app:algorithm}
Algorithm~\ref{alg:ftip} summarizes the minibatch training procedure for
\textsc{FTIP}, including the construction of the finite prior-sample surrogate,
the generation of flow-transformed coefficient samples, and the stochastic
optimization of the resulting variational objective.

\begin{algorithm}[h]
\caption{Training \textsc{FTIP} on a minibatch}
\label{alg:ftip}
\begin{algorithmic}[1]
\Require minibatch $\mathcal B$, prior generator $g(\cdot,\bm \theta)$, number of prior samples $S$, flow posterior $T_{\bm\psi}$, posterior sample count $K$
\State  {Draw fixed $\bm \theta_s\sim p(\bm \theta)$ and form $f_s(\mathbf x_n)=g(\mathbf x_n,\bm \theta_s)$} for $s=1,\ldots,S$; compute
$f_s(\mathbf x_n)=g(\mathbf x_n,\bm \theta_s)$ (noise generator is fixed so sample functions are the same for each minibatch), and compute
$m(\mathbf x_n),\{\phi_s(\mathbf x_n)\}_{s=1}^S$ for $n\in\mathcal B$
\For{$k=1,\dots,K$}
    \State Sample $\bm\varepsilon^{(k)}\sim\mathcal N(\bm 0,\bm I_S)$ and set
    $\bm a^{(k)}=T_{\bm\psi}(\bm\varepsilon^{(k)})$
    \State Form
    $F_n^{(k)}=F(\mathbf x_n;\bm a^{(k)})
    =
    m(\mathbf x_n)+\sum_{s=1}^{S}\phi_s(\mathbf x_n)a_s^{(k)}$
    for $n\in\mathcal B$
\EndFor
\State Compute the loss function in Eq.~\eqref{eq:bbalpha} or the ELBO in Eq.~\eqref{eq:ftip-elbo}
\State Add optional prior-adaptation safeguard $ {R_{\mathrm{prior}}(\bm\theta)}$\State Update trainable parameters with Adam
\end{algorithmic}
\end{algorithm}

\subsection{Warm-start training}
\label{app:warmstart}

We use a warm-start procedure to stabilize optimization of the flow posterior. 
The normalizing flow is initialized close to an affine map, so that the early coefficient posterior has the form
\[
    \bm a
    =
    T_{\mathrm{aff}}(\bm\varepsilon)
    =
    \bm M\bm\varepsilon+\bm b,
    \qquad
    \bm\varepsilon\sim\mathcal N(\bm 0,\bm I_S).
\]
During the first stage of training, only the affine parameters \(\bm M\) and \(\bm b\) are optimized, while the nonlinear spline coupling layers and mixing layers are kept fixed near the identity. This gives a Gaussian coefficient posterior,
\[
    q(\bm a)
    =
    \mathcal N(\bm b,\bm M\bm M^\top),
\]
and therefore a Gaussian predictive distribution under the finite linear surrogate.

 {After the shared affine stage, training branches in two. The \textsc{VIP} branch continues optimizing the affine Gaussian coefficient posterior, while the \textsc{FTIP} branch enables the nonlinear flow layers. Both branches receive the same continuation budget, and all reported matched results are measured after continuation; \textsc{FTIP} is therefore not compared with the earlier checkpoint used to initialize it.}

\begin{table}[t]
\centering
\small
\caption{ {Matched two-branch optimization budgets for \textsc{VIP} and \textsc{FTIP}. The slash in the UCI row denotes the two budgets used across datasets.}}
\label{tab:matched-budgets}
 {
\begin{tabular}{lcccc}
\toprule
Experiment & Shared stage & VIP continuation & FTIP continuation & Total/method \\
\midrule
UCI & $30$k/$60$k & $30$k/$60$k & $30$k/$60$k & $60$k/$120$k \\
Year & $30$k & $30$k & $30$k & $60$k \\
SUSY/HIGGS & $50$k & $50$k & $50$k & $100$k \\
FashionMNIST/CIFAR-10 & $100$ epochs & $100$ epochs & $100$ epochs & $200$ epochs \\
\bottomrule
\end{tabular}}
\end{table}

 {The synthetic diagnostics are the only exception in the submitted configuration: \textsc{FTIP} uses $20\,000$ continuation steps, whereas \textsc{VIP} required $200\,000$ steps to reach convergence. Table~\ref{tab:matched-budgets} shows the full comparison in optimization steps.}

\subsection{Antithetic Sampling}\label{app:antithetic-sampling}

 {To estimate expectations with $K$ coefficient samples, we draw $K/2$ base vectors and pair each with its reflection,}
\begin{equation}
 {
\bm\varepsilon^{(k)}\sim\mathcal N(\bm 0,\bm I_S),
\qquad
\bm\varepsilon^{(k+K/2)}=-\bm\varepsilon^{(k)},
\qquad k=1,\ldots,K/2.}
\end{equation}
 {Each marginal remains a valid base draw, so pushing the pairs through $T_{\bm\psi}$ leaves the Monte Carlo estimator unbiased and does not change the variational family or objective. For a scalar integrand $h$,}
\begin{equation}
 {
\widehat\mu_{\mathrm{ant}}
=
\frac{1}{K}\sum_{k=1}^{K/2}
\left[h(\bm\varepsilon^{(k)})+h(-\bm\varepsilon^{(k)})\right],
\qquad
\operatorname{Var}(\widehat\mu_{\mathrm{ant}})
=
\frac{\sigma^2}{K}(1+\rho),}
\end{equation}
 {where $\rho=\operatorname{Corr}[h(\bm\varepsilon),h(-\bm\varepsilon)]$. Thus antithetic sampling reduces variance only when $\rho<0$; this is not guaranteed after a nonlinear flow and must be checked empirically.}

\begin{table*}[t]
\centering
\scriptsize
\caption{ {Controlled IID-versus-antithetic comparison on the synthetic diagnostics. Predictive metrics are mean $\pm$ standard deviation over matched runs; the last two columns report the variance of the stochastic objective and of the generative-function gradient estimator.}}
\label{tab:antithetic-ablation}
 {
\resizebox{\textwidth}{!}{%
\begin{tabular}{llcccccc}
\toprule
Dataset & Sampling & RMSE & NLL & CRPS & CQM & Objective variance & Generative-gradient variance \\
\midrule
Bimodal & IID
& $10.4404\pm0.1669$
& $2.0897\pm0.0470$
& $5.0362\pm0.1283$
& $0.0297\pm0.0143$
& $17\,218\pm10\,078$
& $316\,303\pm219\,511$ \\
Bimodal & Antithetic
& $10.5719\pm0.2045$
& $2.1016\pm0.0556$
& $5.1567\pm0.1780$
& $0.0346\pm0.0133$
& $7\,273\pm8\,592$
& $275\,058\pm194\,863$ \\
Skewed & IID
& $4.6558\pm0.1368$
& $2.2828\pm0.0229$
& $2.0187\pm0.0622$
& $0.0664\pm0.0022$
& $7\,800\,933\pm436\,997$
& $7\,349\,713\pm1\,115\,122$ \\
Skewed & Antithetic
& $4.8184\pm0.1671$
& $2.2662\pm0.0161$
& $2.0818\pm0.0750$
& $0.0577\pm0.0040$
& $4\,036\,406\pm1\,820\,930$
& $5\,964\,235\pm98\,665$ \\
\bottomrule
\end{tabular}}}
\end{table*}

 {To isolate the estimator from learned parameters, IID and antithetic estimates were evaluated at identical parameter states (Table~\ref{tab:antithetic-ablation}). Each of $500$ estimates used the same fixed seed-42 minibatch and a matched replicate seed. Across common trained checkpoints, antithetic sampling reduced stochastic-objective variance by $34$--$88\%$ and also reduced generative-function gradient variance. We therefore describe it only as an empirical training-stability device, not as a source of posterior expressiveness or an unmatched advantage over \textsc{VIP}.}

\section{Experimental Details}
\label{app:experimental-details}

All experiments use the Adam optimiser with an initial learning rate of $10^{-3}$ and cosine annealing down to $\eta_{\min} = 10^{-5}$.
Unless noted otherwise, models are trained in double precision (\texttt{float64}).
Antithetic sampling is used throughout to reduce Monte Carlo variance, as described in Appendix~\ref{app:antithetic-sampling}.

All training times are reported averaging $10\ 000$ iterations on a Nvidia 3060 Ti. 

\paragraph{ {Implicit BNN prior.}}\label{app:bnn-prior}
 {For each prior-function draw $s$ and layer $\ell$, weights and biases are sampled independently as}
\begin{equation}
 {
\bm W_{\ell}^{(s)}
=
\bm\mu_{W_\ell}
+
\bm\sigma_{W_\ell}\odot\bm\epsilon_{\ell}^{(s)},
\qquad
\bm b_{\ell}^{(s)}
=
\bm\mu_{b_\ell}
+
\bm\sigma_{b_\ell}\odot\bm\xi_{\ell}^{(s)},
\qquad
\bm\epsilon_{\ell}^{(s)},\bm\xi_{\ell}^{(s)}
\sim
\mathcal N(\bm 0,\bm I).}
\end{equation}
 {The initial implicit prior has zero means and unit standard deviations. We denote the collection of layerwise means and standard deviations by $\bm\lambda$ (four scalars per layer, broadcast across the corresponding weight and bias entries, following \cite{ortega2023deep}); in the regression experiments these parameters may be adapted using the empirical-Bayes procedure inherited from \textsc{VIP}. This weight-space prior is distinct from the surrogate coefficient prior $p(\bm a)=\mathcal N(\bm 0,\bm I_S)$ and the coefficient posterior $q_{\bm\psi}(\bm a)$.}

\paragraph{Flow architecture.}
For all \textsc{FTIP} experiments the normalizing flow $T_{\bm\psi}$ is an initial learnable affine map followed by $L=2$ rational-quadratic spline coupling layers interleaved with $1\times 1$ LU mixing layers (spline$+$1x1), as described in Section~\ref{sec:ftip}.
Each spline has $R=8$ bins and is defined on the domain $[-B,B]$ with $B=3.0$.
The hidden width of the coupling-layer MLPs is twice the input dimension.
This warm-start initialization means the affine map already represents a Gaussian posterior before the nonlinear layers are enabled.

\subsection{Synthetic regression diagnostics}

The synthetic diagnostics are designed to isolate posterior-predictive effects that are difficult for Gaussian or effectively unimodal approximations. We use two one-dimensional regression problems: one with a bimodal conditional distribution and one with asymmetric, input-dependent skewness. These experiments are not intended primarily as point-prediction benchmarks; instead, they test whether the posterior predictive distribution can place probability mass on the appropriate parts of the conditional target distribution.

\paragraph{Bimodal dataset.}
Inputs are sampled uniformly as
\[
    x \sim \mathrm{Unif}[-4,4].
\]
Targets are generated from one of two equally likely sinusoidal branches,
\[
    y \mid x
    \sim
    \frac{1}{2}
    \mathcal N\!\left(20\cos(x-0.5),1\right)
    +
    \frac{1}{2}
    \mathcal N\!\left(20\sin(x-0.5),1\right).
\]
Equivalently,
\[
    y =
    \begin{cases}
    20\cos(x-0.5)+\epsilon, & b=0,\\
    20\sin(x-0.5)+\epsilon, & b=1,
    \end{cases}
    \qquad
    b\sim\mathrm{Bernoulli}(1/2),
    \qquad
    \epsilon\sim\mathcal N(0,1).
\]
For many values of $x$, the two conditional branches are well separated. A Gaussian predictive approximation must therefore either average the two branches or use a large variance to cover both. This makes the dataset a direct diagnostic for multimodal predictive uncertainty.

\paragraph{Skewed dataset.}
The second diagnostic uses asymmetric conditional noise. Inputs are again sampled from
\[
    x \sim \mathrm{Unif}[-4,4],
\]
and targets are generated as
\[
    y = 5\sin(x) + s(x)\xi,
    \qquad
    s(x)=
    \begin{cases}
    -1, & x<0,\\
    1, & x\geq 0.
    \end{cases}
\]
Here $\xi$ is a centered log-normal perturbation,
\[
    \xi = \exp(\sigma z)-\exp(\sigma^2/2),
    \qquad
    z\sim\mathcal N(0,1),
    \qquad
    \sigma=0.8.
\]
The sign factor $s(x)$ reverses the direction of the tail at the origin. Thus, the conditional distribution is left-skewed for $x<0$ and right-skewed for $x\geq0$. A symmetric predictive approximation must over-disperse to cover the heavy tail, whereas a more flexible posterior can represent the asymmetric uncertainty directly.

\paragraph{Architecture.}
All methods use a two-hidden-layer Bayesian neural network with widths $[10,10]$ and $\tanh$ activations.
For \textsc{VIP} and \textsc{FTIP} the prior is represented by $S=20$ surrogate samples.

\paragraph{Hyperparameters.}
\textsc{VIP} is trained with $\alpha=1.0$, learning rate $10^{-4}$, batch size $200$, and $200\,000$ iterations.
 {\textsc{FTIP} is initialized from the converged \textsc{VIP} checkpoint and trained for $20\,000$ iterations with learning rate $10^{-3}$ and $\alpha=1.0$. This is the only experiment in which the submitted configurations use unequal optimization budgets.}
\textsc{MFVI} uses a diagonal Gaussian posterior over the full parameter vector, $\alpha=0.5$, learning rate $10^{-3}$, and $200\,000$ iterations.
\textsc{FBNN} uses $20$ measurement points and $20$ context points with context standard deviation $2.0$, a GP prior, $\lambda_{\mathrm{KL}}=1.0$, $\alpha=1.0$, learning rate $10^{-3}$, and $200\,000$ iterations.
\textsc{TFSVI} uses $S_{\mathrm{ctx}}=5$ context functions, $K_{\mathrm{ctx}}=20$ context points, $\sigma_{\mathrm{prior}}=1.0$, $\alpha=1.0$, learning rate $10^{-3}$, and $200\,000$ iterations.
All methods draw $K=20$ Monte Carlo samples per gradient step during training and $100$ samples at evaluation.

\subsection{UCI regression benchmarks}

\paragraph{Datasets and splits.}
We use the standard UCI regression benchmark comprising nine datasets: Boston, Concrete, Energy, Kin8nm, Naval, Power, Protein, Wine, and Yacht.
Each dataset is split into a $90\%$ training set and a $10\%$ held-out test set.
Results are reported as mean and standard deviation over $5$ independent random seeds.

\paragraph{Architecture.}
All methods use a two-hidden-layer Bayesian neural network with widths $[10,10]$, $\tanh$ activations, and the a parameterization which shares scalar mean and variance parameters across all weights and biases within each layer.
For \textsc{VIP} and \textsc{FTIP} the prior is represented by $S=20$ surrogate samples.

\paragraph{Hyperparameters.}
 {\textsc{VIP} and \textsc{FTIP} use the matched two-branch protocol in Table~\ref{tab:matched-budgets}: they share a $30\,000$- or $60\,000$-step affine stage, depending on the dataset, and each receives an equal continuation budget. The reported values are taken after both branches, for equal totals of $60\,000$ or $120\,000$ updates per method; the nonlinear \textsc{FTIP} continuation uses learning rate $10^{-4}$.}
\textsc{MFVI} uses $300\,000$ iterations and $\alpha=0.5$.
\textsc{FBNN} and \textsc{TFSVI} use the same settings as in the synthetic experiments.
We evaluate both $\alpha=0.5$ and $\alpha=1.0$ for \textsc{VIP} and \textsc{FTIP}; Table~\ref{tab:uci_simpler_a05_nll} reports results for $\alpha=0.5$.
Timing results in Table~\ref{tab:uci_simpler_a05_nll} are median wall-clock milliseconds per iteration measured on a single GPU.

\subsection{ YearPredictionMSD}
 {For the YearPredictionMSD results, \textsc{VIP} and \textsc{FTIP} share a $30\,000$-step affine stage and then receive $30\,000$ continuation updates each, for $60\,000$ updates per reported method.}

\subsection{Pedestrian future path-length prediction}

\paragraph{Dataset and splits.}
We use the ETH/UCY pedestrian benchmark, pooling trajectories from eight scenes.
The dataset is split into $24\,264$ training, $5\,031$ validation, and $7\,975$ test trajectories, with no pedestrian shared across splits.
Each trajectory provides $T_{\mathrm{obs}}=8$ observed positions and $T_{\mathrm{pred}}=12$ future positions; the regression target is the scalar total Euclidean path length of the $12$ future steps.

\paragraph{Architecture.}
All methods share a BayesianLSTM backbone: a deterministic single-layer LSTM encoder with hidden size $64$, followed by a Bayesian MLP head with widths $[64, 32]$ and a scalar output.
For \textsc{VIP} and \textsc{FTIP} the surrogate coefficient dimension is $S=40$.
\textsc{MFVI}, \textsc{FBNN}, and \textsc{TFSVI} place a diagonal Gaussian posterior over the full approximately $17\,000$-dimensional parameter vector.

\paragraph{Hyperparameters.}
All methods are trained for $30\,000$ iterations with batch size $128$, Adam, learning rate $10^{-3}$ with cosine annealing, and $\alpha=1.0$.
\textsc{FTIP} uses the warm-start protocol with a two-layer spline$+$1x1 flow.
Predictive metrics are evaluated with $K=1\,000$ posterior samples.

\subsection{Model correction under prior misspecification}

\paragraph{Setup.}
We use a Gaussian process with a squared-exponential kernel as the base prior, parameterized by amplitude and length scale, both initialised to $1.0$.
The prior is represented by $S=30$ exact GP samples.
Kernel hyperparameters are optimised by marginal log-likelihood maximisation for $2\,000$ gradient steps.
\textsc{FTIP} is then applied with a two-layer spline$+$1x1 flow on the bimodal regression dataset described above.

\subsection{Image classification}

\paragraph{Datasets.}
We evaluate on two image classification benchmarks: FashionMNIST (same dimensions and splits) and CIFAR-10 ($50\,000$ train / $10\,000$ test, $32{\times}32$ RGB, 10 classes).
No data augmentation is applied.
Results are reported as mean and standard deviation over 5 independent random seeds.

\paragraph{Architecture.}
All methods share a LeNet-style backbone: two deterministic convolutional layers (6 and 16 filters, $5{\times}5$ kernels) followed by a Bayesian MLP head with widths $[120, 84]$ and \texttt{ReLU} activations.
The convolutional feature extractor is shared and deterministic; randomness enters only through the Bayesian head.
For \textsc{VIP} and \textsc{FTIP} the prior is represented by $S=20$ surrogate samples, with weight log-standard-deviation initialised at $0.0$ (i.e.\ $\sigma_w=1.0$).
For \textsc{MFVI}, \textsc{FBNN}, and \textsc{TFSVI} the weight log-standard-deviation is initialised at $-3.0$ ($\sigma_w \approx 0.05$) to avoid posterior collapse during early training.
All experiments use \texttt{float32} precision and batch size $256$.

\paragraph{Hyperparameters.}
 {\textsc{MFVI}, \textsc{FBNN}, and \textsc{TFSVI} are trained for $100$ epochs with Adam and cosine annealing. \textsc{VIP} and \textsc{FTIP} share the first $100$ epochs and then receive $100$ continuation epochs each; the reported \textsc{VIP} and \textsc{FTIP} values are measured after $200$ epochs per method. The \textsc{VIP} branch continues optimizing its affine Gaussian posterior, while \textsc{FTIP} enables the nonlinear flow. Both use $\alpha=0.5$ and no prior regularizer. On FashionMNIST the \textsc{FTIP} continuation uses learning rate $10^{-4}$ with a depth-$4$ spline$+$1$\times$1 flow; on CIFAR-10 it uses $10^{-5}$ and depth $2$.}
All spline layers use $K=8$ bins on domain $[-3, 3]$.
\textsc{MFVI} uses $\alpha=0.5$, $200$ weight samples at evaluation.
\textsc{FBNN} uses $20$ measurement points, $20$ context points with context standard deviation $2.0$, a GP prior, $\lambda_{\mathrm{KL}}=1.0$, and $\alpha=0.5$.
\textsc{TFSVI} uses $S_{\mathrm{ctx}}=5$, $K_{\mathrm{ctx}}=20$, $\sigma_{\mathrm{prior}}=1.0$, and $\alpha=0.5$.
Predictive probabilities for all methods are computed via $1\,000$ Monte Carlo samples at test time.

\subsection{Binary classification}

\paragraph{Datasets.}
We evaluate on two large-scale binary classification benchmarks: HIGGS ($10\,500\,000$ samples, $28$ kinematic features, signal/background separation) and SUSY ($5\,000\,000$ samples, $18$ kinematic features).
Both datasets are used in their standard train/test splits.
To keep per-step evaluation tractable, all intermediate and final metrics are computed on a random subsample of $50\,000$ examples per split.
Results are reported as mean and standard deviation over 5 independent random seeds.
No \textsc{CNF} baseline is included, as the CNF classifier is multiclass-only.

\paragraph{Architecture.}
All methods use a two-hidden-layer Bayesian MLP with widths $[100, 100]$, \texttt{ReLU} activations with a scalar output and Bernoulli/probit likelihood.
For \textsc{VIP} and \textsc{FTIP} the weight log-standard-deviation is initialised at $0.0$ ($\sigma_w = 1.0$).
For \textsc{MFVI}, \textsc{FBNN}, and \textsc{TFSVI} it is initialised at $-1.0$ ($\sigma_w \approx 0.37$) to prevent logit saturation under the probit link at the start of training.
All methods use $S=20$ surrogate samples and $\alpha=0.5$.

\paragraph{Hyperparameters.}
 {\textsc{VIP} and \textsc{FTIP} share a $50\,000$-iteration affine stage and then receive $50\,000$ continuation iterations each, for $100\,000$ updates per reported method. Training uses batch size $1\,024$, Adam, and cosine annealing; the nonlinear \textsc{FTIP} continuation uses learning rate $10^{-4}$ and a depth-$4$ spline$+$1$\times$1 flow with $K=8$ bins on $[-3,3]$.}
Baseline-specific settings follow those of the multiclass experiments: \textsc{FBNN} uses $20$ measurement points, $20$ context points, context standard deviation $2.0$, and $\lambda_{\mathrm{KL}}=1.0$; \textsc{TFSVI} uses $S_{\mathrm{ctx}}=5$, $K_{\mathrm{ctx}}=20$, and $\sigma_{\mathrm{prior}}=1.0$.
Predictive probabilities are obtained by passing $200$ posterior samples through the probit link and averaging.

\section{Further Results and Experiments}
This appendix reports additional experimental results that complement the main
text. We include the full UCI regression tables, additional classification
experiments, and a short discussion of the relationship between \textsc{CRPS}
and \textsc{RMSE}. These results are intended to make the empirical comparison
more complete and to clarify when the additional expressiveness of
\textsc{FTIP} is beneficial.

\subsection{ HMC and deep-ensemble diagnostics}
\label{app:broader-baselines}

 {We compare the synthetic diagnostics with HMC~\citep{neal1996bayesian} and a five-member deep ensemble, using the same $1$--$10$--$10$--$1$ BNN, Gaussian weight prior, and Gaussian observation family as the variational baselines. Table~\ref{tab:broader-synthetic} reports averages over five random seeds.}

\begin{table*}[t]
\centering
\scriptsize
\caption{ {Five-seed synthetic diagnostics for \textsc{VIP}, \textsc{FTIP}, HMC, and a standard deep ensemble.}}
\label{tab:broader-synthetic}
 {
\resizebox{\textwidth}{!}{%
\begin{tabular}{llccccc}
\toprule
Dataset & Method & RMSE $\downarrow$ & NLL $\downarrow$ & CRPS $\downarrow$ & CQM $\downarrow$ & Predictive structure \\
\midrule
\multirow{4}{*}{Bimodal}
& VIP
& $10.45\pm0.09$
& $3.49\pm0.01$
& $5.76\pm0.06$
& $0.222\pm0.011$
& Unimodal/symmetric \\
& FTIP
& $10.28\pm0.20$
& $2.10\pm0.06$
& $4.92\pm0.18$
& $0.017\pm0.013$
& Recovers bimodality \\
& HMC
& $10.45\pm0.09$
& $3.77\pm0.01$
& $6.17\pm0.06$
& $0.120\pm0.011$
& Broad/unimodal \\
& Deep ensemble
& $10.48\pm0.09$
& $3.77\pm0.01$
& $6.18\pm0.05$
& $0.120\pm0.011$
& Near-unimodal \\
\midrule
\multirow{4}{*}{Skewed}
& VIP
& $3.95\pm0.15$
& $2.80\pm0.04$
& $1.93\pm0.05$
& $0.118\pm0.003$
& Approximately symmetric \\
& FTIP
& $4.45\pm0.17$
& $2.24\pm0.02$
& $1.91\pm0.08$
& $0.062\pm0.004$
& Recovers asymmetry \\
& HMC
& $3.94\pm0.15$
& $2.79\pm0.04$
& $1.93\pm0.05$
& $0.118\pm0.003$
& Approximately symmetric \\
& Deep ensemble
& $3.94\pm0.15$
& $2.79\pm0.04$
& $1.92\pm0.05$
& $0.118\pm0.003$
& Near-unimodal \\
\bottomrule
\end{tabular}}}
\end{table*}

 {HMC uses four chains with $500$ discarded and $1\,000$ retained draws per chain, $500$ leapfrog steps, and step size $2.5\times10^{-3}$. Across $400$ fixed-grid function values and the log-noise parameter, all tasks satisfy $\widehat R\leq1.004$, minimum bulk/tail effective sample sizes above $1\,700/2\,400$, and zero divergences. The broad predictions therefore do not appear to be caused by an obvious chain-convergence failure.}

 {The comparison must nevertheless separate posterior family from inferential target. HMC targets the ordinary posterior}
\begin{equation}
 {
p(\bm w,\sigma\mid\mathcal D)
\propto
p(\bm w,\sigma)
\prod_i p(y_i\mid\bm x_i,\bm w,\sigma),}
\end{equation}
 {under which every weight draw must explain the complete dataset. By contrast, \textsc{FTIP} with $\alpha=1$ optimizes the per-observation predictive mixture}
\begin{equation}
 {
\sum_i
\log
\left[
\frac{1}{K}
\sum_{k=1}^{K}
p\!\left(y_i\mid F^{(k)}(\bm x_i),\sigma\right)
\right],}
\end{equation}
 {so different function samples may specialize across branches or response regions. \textsc{VIP} uses the same objective but retains a Gaussian coefficient posterior. A standard deep ensemble trains its members independently, so each member is likewise driven to explain all observations; a jointly trained mixture ensemble would be a different objective and model class. These diagnostics demonstrate greater posterior-predictive flexibility in the stated setting, not better approximation of the ordinary HMC posterior or universal superiority over MCMC, SGLD, SWAG, or alternative ensemble objectives.}

\subsection{Full UCI results}

Table~\ref{tab:uci_simpler_full} reports the complete UCI regression results
for all datasets, metrics, and values of $\alpha$ considered in the experiments.
The main text focuses on representative negative log-likelihood results, while
the full table also includes \textsc{RMSE}, \textsc{CRPS}, and CQM. This gives a
more complete view of the trade-offs between point-prediction accuracy,
distributional sharpness, and calibration-sensitive metrics.

Across the full set of UCI benchmarks, \textsc{FTIP} is generally competitive
with \textsc{VIP}. On datasets where the posterior predictive distribution is
well described by a unimodal Gaussian-like approximation, the flow posterior
does not systematically improve over the Gaussian coefficient posterior used by
\textsc{VIP}. This is expected: in such cases, the additional flexibility of the
flow is not strongly used by the objective. On the other hand, on datasets where
the predictive distribution appears more structured, \textsc{FTIP} can improve
likelihood-based and calibration-sensitive metrics, especially under the more
mass-covering setting $\alpha=1.0$. These trends support the interpretation of
\textsc{FTIP} as an expressive extension of \textsc{VIP} rather than as a
uniform replacement for the Gaussian surrogate posterior.

\subsection{Posterior correction under prior misspecification}
\label{app:prior-misspecification}

\begin{figure}[t]
    \centering
    \includegraphics[width=1\linewidth]{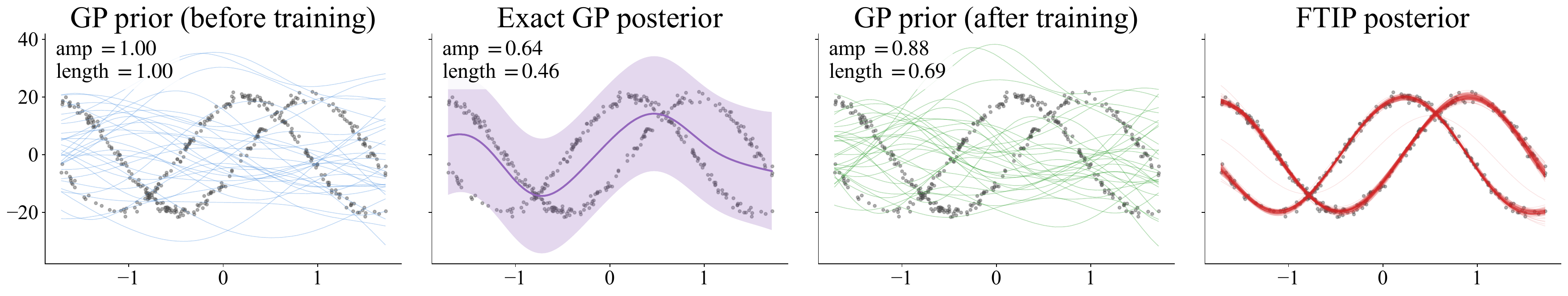}
    \caption{Posterior correction under GP prior misspecification.
    \textbf{First:} samples from the initial GP prior.
    \textbf{Second:} exact GP posterior after fitting the kernel
    hyperparameters.
    \textbf{Third:} samples from the GP prior after hyperparameter adaptation in
    \textsc{FTIP}.
    \textbf{Fourth:} posterior samples from \textsc{FTIP}.}
    \label{fig:gp_ftip_training}
\end{figure}

We include an additional diagnostic to test whether \textsc{FTIP} can partially
compensate for a misspecified base prior. The task is the bimodal regression
problem described in Appendix~\ref{app:experimental-details}. As a deliberately
restrictive prior, we use a Gaussian process with a squared-exponential kernel.
Even after optimizing the GP kernel hyperparameters, the exact GP posterior
remains Gaussian. Consequently, it cannot represent two separated posterior
functional explanations; instead, it tends to average across the two branches
and inflate uncertainty between them.

Figure~\ref{fig:gp_ftip_training} illustrates this effect. The first panel shows
samples from the initial GP prior, and the second panel shows the exact GP
posterior after hyperparameter fitting. Although the fitted GP adapts its length
scale and marginal variance to the data, its posterior is still Gaussian in
function space, so the resulting samples do not cleanly separate the two
predictive branches. The third panel shows samples from the adapted GP prior
used by \textsc{FTIP}. The prior itself is still Gaussian and therefore remains
misspecified for this bimodal task.

The fourth panel shows posterior samples from \textsc{FTIP}. In this case, the
finite GP prior-sample surrogate is combined with a flow-transformed coefficient
posterior. The flow posterior induces a non-Gaussian distribution over surrogate
functions, allowing different posterior samples to occupy different branches of
the bimodal target distribution. Thus, \textsc{FTIP} does not make the GP prior
itself multimodal, but it can use a more expressive posterior over the sampled
function basis to represent multimodal posterior structure. This suggests that
posterior expressiveness can partly mitigate prior misspecification, although it
does not remove the need for a sufficiently rich prior-sample basis.

\subsection{Ablation Studies}
We ablate four hyperparameters of FTIP on the \textsc{Year} regression task in a
one-at-a-time fashion, varying a single axis while holding the rest at the
default configuration ($S=20$ regression coefficients, $K=200$ Monte~Carlo
samples per training step, $\alpha=1.0$ for the BB-$\alpha$ energy, and a
flow of depth $2$). Each configuration is trained for 60\,000 iterations under
the auto warm-start protocol and repeated across 5 random seeds. Table~\ref{tab:year_ablation}
reports test RMSE, NLL, CRPS, and CQM averaged over the 5 seeds; for the
FTIP-only knobs ($K$ and depth) the matching VIP baseline is identical for
every value of the axis and is therefore reported once per block.

\paragraph{Number of regression coefficients $S$.}
The size of the basis is by far the most influential hyperparameter: increasing
$S$ from $5$ to $100$ lowers FTIP's RMSE from $10.76$ to $9.72$, NLL from
$3.46$ to $3.29$, and CRPS from $5.50$ to $4.76$, with VIP showing a parallel
trend. This is consistent with the implicit-process interpretation of the
model: $S$ controls the dimensionality of the function-space basis the flow
posterior is defined over, so a larger $S$ increases the expressive power of
the family. The improvement is monotonic on every metric and shows no sign
of saturating at $S=100$, suggesting that further gains are still attainable
at the cost of additional compute (the cubic Cholesky factor in the closed-form
predictive scales as $S^3$, so we cap $S$ at $100$ in practice).  {We treat $S$ as a validation-selected basis-capacity parameter and always use the same value for \textsc{VIP} and \textsc{FTIP}; the standard experiments use the \textsc{VIP} default $S=20$, increased to $S=40$ only for the pedestrian task.}

\paragraph{Monte~Carlo budget $K$.}
The number of posterior samples used during training has only a marginal
effect on FTIP and none on VIP (whose predictive is closed-form). Going from
$K=50$ to $K=1000$ gains FTIP $\sim 0.14$ RMSE and $0.03$ NLL --- a small
improvement reflecting variance reduction in the gradient estimator
$\nabla_\theta \widehat{\mathcal{L}}$, but not a structural one. In particular,
$K=200$ (our default) is essentially indistinguishable from $K=500$ or
$K=1000$ on every metric within seed variability, so there is little to
gain from inflating the per-step Monte~Carlo budget beyond a few hundred
samples on this scale of dataset.

\paragraph{BB-$\alpha$ energy $\alpha$.}
The choice of $\alpha$ controls a smooth interpolation between the standard
ELBO ($\alpha=0$) and the BB-$\alpha$ energy at $\alpha=1$. At $\alpha=0$ and $\alpha=0.5$ the model achieves slightly
lower RMSE ($\sim 10.18$--$10.23$) but substantially worse calibration: NLL
sits at $3.74$, CRPS at $5.47$--$5.49$, and the centered-quantile metric
CQM at $0.05$. Switching to $\alpha=1.0$ trades $0.1$ RMSE for a markedly
sharper and better-calibrated predictive distribution: NLL drops to $3.41$,
CRPS to $5.22$, and CQM by an order of magnitude to $0.007$. The same qualitative pattern shows up in VIP.  {This ablation illustrates the objective trade-off rather than defining a test-set selection rule. In practice, $\alpha$ is selected on a validation set using the metric relevant to the application, and the selected value is then shared by \textsc{VIP} and \textsc{FTIP}.}

\paragraph{Flow depth.}
Increasing the number of coupling layers from $1$ to $10$ produces only modest
gains: RMSE goes from $10.48$ to $10.30$ and NLL from $3.43$ to $3.40$, with
performance plateauing around depth $4$--$6$. Deeper flows do not destabilise
training in our setup, but the marginal returns past depth $4$ are small,
suggesting that the regression-coefficient posterior on \textsc{Year} is well
captured by a relatively shallow flow. The default depth of $2$ already
recovers most of the benefit and is preferred for its lower compute cost.

\begin{table*}[t]
    \centering
    \small
    \caption{Ablation study on YearPredictionMSD. Each block varies one design choice while
holding the remaining configuration fixed. For design choices that affect only
\textsc{FTIP}, the corresponding \textsc{VIP} baseline is unchanged and is shown
once per block. Best results are highlighted in teal. Time denotes wall-clock
training time in minutes on a single GPU. Values are medians over 6 seeds.}
    \label{tab:year_ablation}
    \scalebox{0.8}{
    \begin{tabular}{llcccccccccc}
        \toprule
        Axis & Value & \multicolumn{5}{c}{FTIP} & \multicolumn{5}{c}{VIP} \\
        \cmidrule(lr){3-7} \cmidrule(lr){8-12}
          &       & RMSE $\downarrow$ & NLL $\downarrow$ & CRPS $\downarrow$ & CQM $\downarrow$ & Time & RMSE $\downarrow$ & NLL $\downarrow$ & CRPS $\downarrow$ & CQM $\downarrow$ & Time \\
        \midrule
        \multirow{5}{*}{$S$}
         & $5$   & $10.76$ & $3.46$ & $5.50$ & $0.010$ & $17.7$ & $10.62$ & $3.75$ & $5.64$ & $0.039$ & $14.3$ \\
         & $10$  & $10.62$ & $3.44$ & $5.37$ & $0.008$ & $17.7$ & $10.49$ & $3.72$ & $5.52$ & $0.035$ & $14.2$ \\
         & $20$  & $10.30$ & $3.41$ & $5.22$ & $0.007$ & $17.9$ & $10.24$ & $3.66$ & $5.34$ & $0.029$ & $14.2$ \\
         & $40$  & $10.18$ & $3.37$ & $5.09$ & $0.009$ & $18.3$ & $10.08$ & $3.61$ & $5.19$ & $0.026$ & $14.5$ \\
         & $100$ & \bm{$9.72$} & \bm{$3.29$} & \bm{$4.76$} & \bm{$0.006$} & $19.5$ & \bm{$9.60$} & \bm{$3.50$} & \bm{$4.81$} & \bm{$0.021$} & $14.4$ \\
        \midrule
        \multirow{5}{*}{$K$}
         & $50$   & $10.41$ & $3.43$ & $5.27$ & $0.013$ & $18.9$ \\
         & $100$  & $10.35$ & $3.42$ & $5.23$ & $0.011$ & $17.5$ & & & & & \\
         & $200$  & $10.31$ & $3.41$ & $5.22$ & $0.009$ & $17.9$ & & & & & \\
         & $500$  & $10.28$ & \bm{$3.40$} & $5.21$ & \bm{$0.004$} & $18.7$ & & & & & \\
         & $1000$ & \bm{$10.27$} & \bm{$3.40$} & \bm{$5.20$} & $0.007$ & $24.2$ & & & & & \\
        \midrule
        \multirow{3}{*}{$\alpha$}
         & $0.0$ & $10.23$ & $3.74$ & $5.49$ & $0.050$ & $18.5$ & $10.24$ & $3.74$ & $5.50$ & $0.049$ & $15.5$ \\
         & $0.5$ & \bm{$10.19$} & $3.74$ & $5.47$ & $0.051$ & $17.7$ & \bm{$10.19$} & $3.74$ & $5.48$ & $0.050$ & $14.1$ \\
         & $1.0$ & $10.30$ & \bm{$3.41$} & \bm{$5.22$} & \bm{$0.007$} & $21.9$ & $10.24$ & \bm{$3.66$} & \bm{$5.34$} & \bm{$0.029$} & $18.4$ \\
        \midrule
        \multirow{5}{*}{depth}
         & $1$  & $10.48$ & $3.43$ & $5.31$ & $0.009$ & $12.2$ \\ 
         & $2$  & $10.31$ & $3.41$ & $5.22$ & $0.008$ & $17.8$ & & & & & \\
         & $4$  & $10.32$ & \bm{$3.40$} & $5.21$ & $0.010$ & $28.2$ & & & & & \\
         & $6$  & $10.31$ & \bm{$3.40$} & \bm{$5.20$} & \bm{$0.006$} & $40.1$ & & & & & \\
         & $10$ & \bm{$10.30$} & \bm{$3.40$} & $5.21$ & $0.008$ & $60.7$ & & & & & \\
        \bottomrule
    \end{tabular}}
\end{table*}

\subsection{ Computational profiling}
\label{app:computational-profiling}

\begin{table*}[t]
\centering
\scriptsize
\caption{ {Computational cost of the matched \textsc{VIP} and \textsc{FTIP} configurations. Latency is measured per example for one and $1\,000$ posterior samples.}}
\label{tab:computational-profiling}
 {
\resizebox{\textwidth}{!}{%
\begin{tabular}{llrrrrr}
\toprule
Dataset/configuration
& Method
& Trainable params
& Peak memory
& Training time
& $K=1$ latency
& $K=1000$ latency \\
\midrule
Year, $S=20$, depth $2$
& VIP
& $2\,293$
& $16.9$ MiB
& $18.16$ min
& $0.002776$ ms
& $0.002776$ ms \\
Year, $S=20$, depth $2$
& FTIP
& $23\,673$
& $20.4$ MiB
& $37.85$ min
& $0.016354$ ms
& $0.017029$ ms \\
HIGGS, $S=20$, depth $4$
& VIP
& $26\,432$
& $50.4$ MiB
& $10.88$ min
& $0.000376$ ms
& $0.000376$ ms \\
HIGGS, $S=20$, depth $4$
& FTIP
& $69\,192$
& $52.2$ MiB
& $50.18$ min
& $0.003667$ ms
& $0.003920$ ms \\
\bottomrule
\end{tabular}}}
\end{table*}

 {Table~\ref{tab:computational-profiling} shows the results of the computational profiling. For \textsc{FTIP}, each latency measurement freshly samples coefficient vectors, evaluates the basis once per fixed batch, and materializes predictions through a vectorized projection; this shared basis evaluation explains the weak dependence on $K$, whereas \textsc{VIP} prediction is analytic. Increasing $S$ from $5$ to $100$ raises the number of parameters from $3\,771$ to $530\,113$, peak memory from $18.63$ to $37.16$ MiB, and iteration time from $29.014$ to $30.908$ ms, while NLL improves from $3.46$ to $3.29$. Increasing flow depth from $1$ to $10$ raises parameters from $12\,983$ to $109\,193$, memory from $19.33$ to $28.94$ MiB, and iteration time from $21.803$ to $145.328$ ms; NLL saturates around depth $4$--$6$.}

\begin{table}[p]
\centering
\scriptsize
\setlength{\tabcolsep}{2.5pt}
\renewcommand{\arraystretch}{0.92}
\caption{Full UCI regression results across all methods, objectives, datasets, and
evaluation metrics. {\color{teal}\textbf{Best}} and {\color{purple}\textbf{second-to-best}} results are highlighted for each
dataset, value of $\alpha$, and metric. All metrics are lower is better. Results are averaged over $5$ different random seeds. }
\label{tab:uci_simpler_full}
\resizebox{\textwidth}{!}{%
\begin{tabular}{llccccccccc}
\toprule
Metric & Method
& Boston & Concrete & Energy & Kin8nm & Naval & Power & Protein & Wine & Yacht \\
\midrule
\multicolumn{11}{l}{$\alpha=0.5$} \\
\midrule
\multirow{5}{*}{RMSE} & VIP & \second{$3.824$} & \best{$7.975$} & \best{$2.991$} & \best{$0.180$} & \second{$0.005$} & \second{$4.066$} & \best{$4.961$} & \second{$0.637$} & \best{$4.366$} \\
 & FTIP & \best{$3.753$} & \second{$8.032$} & \second{$3.001$} & \second{$0.182$} & \best{$0.005$} & \best{$4.016$} & \second{$4.963$} & \best{$0.637$} & \second{$4.830$} \\
 & FBNN & $7.741$ & $14.099$ & $9.197$ & $0.253$ & $0.015$ & $13.509$ & $6.102$ & $0.750$ & $12.884$ \\
 & MFVI & $9.035$ & $16.408$ & $9.709$ & $0.254$ & $0.015$ & $13.330$ & $6.100$ & $0.748$ & $13.302$ \\
 & TFSVI & $4.996$ & $12.026$ & $4.164$ & $0.234$ & $0.012$ & $4.541$ & $5.695$ & $0.672$ & $9.224$ \\
\cmidrule(lr){1-11}
\multirow{5}{*}{NLL} & VIP & \second{$2.65$} & \best{$3.48$} & \second{$2.11$} & \best{$-0.30$} & \best{$-4.48$} & \best{$2.82$} & \best{$3.02$} & \second{$0.96$} & \second{$2.58$} \\
 & FTIP & \best{$2.64$} & \second{$3.48$} & \best{$2.03$} & \second{$-0.28$} & \second{$-4.23$} & \second{$2.83$} & \second{$3.02$} & \best{$0.95$} & \best{$2.49$} \\
 & FBNN & $3.46$ & $4.07$ & $3.64$ & $0.05$ & $-2.80$ & $4.03$ & $3.23$ & $1.14$ & $3.83$ \\
 & MFVI & $3.62$ & $4.22$ & $3.70$ & $0.05$ & $-2.80$ & $4.00$ & $3.23$ & $1.14$ & $4.04$ \\
 & TFSVI & $3.04$ & $3.91$ & $2.85$ & $-0.04$ & $-2.97$ & $2.93$ & $3.16$ & $1.03$ & $3.65$ \\
\cmidrule(lr){1-11}
\multirow{5}{*}{CRPS} & VIP & \best{$2.015$} & \best{$4.414$} & \second{$1.424$} & \best{$0.101$} & \best{$0.002$} & \best{$2.262$} & \second{$2.834$} & \second{$0.354$} & \best{$2.173$} \\
 & FTIP & \second{$2.182$} & \second{$4.448$} & \best{$1.380$} & \second{$0.102$} & \second{$0.002$} & \second{$2.288$} & \best{$2.818$} & \best{$0.348$} & \second{$2.209$} \\
 & FBNN & $4.514$ & $7.939$ & $5.066$ & $0.145$ & $0.009$ & $7.626$ & $3.579$ & $0.424$ & $6.927$ \\
 & MFVI & $4.942$ & $9.361$ & $5.657$ & $0.146$ & $0.009$ & $7.609$ & $3.564$ & $0.421$ & $7.404$ \\
 & TFSVI & $2.783$ & $6.681$ & $2.304$ & $0.133$ & $0.007$ & $2.534$ & $3.288$ & $0.378$ & $5.175$ \\
\cmidrule(lr){1-11}
\multirow{5}{*}{CQM} & VIP & \second{$0.046$} & $0.031$ & \second{$0.035$} & $0.014$ & \second{$0.040$} & $0.016$ & \best{$0.020$} & \best{$0.025$} & \best{$0.047$} \\
 & FTIP & \best{$0.043$} & \second{$0.028$} & \best{$0.033$} & \best{$0.013$} & $0.043$ & $0.015$ & \second{$0.020$} & \second{$0.026$} & \second{$0.048$} \\
 & FBNN & $0.062$ & $0.042$ & $0.101$ & $0.017$ & $0.074$ & \second{$0.010$} & $0.110$ & $0.095$ & $0.083$ \\
 & MFVI & $0.068$ & $0.034$ & $0.096$ & $0.015$ & $0.072$ & \best{$0.010$} & $0.113$ & $0.110$ & $0.107$ \\
 & TFSVI & $0.065$ & \best{$0.018$} & $0.059$ & \second{$0.014$} & \best{$0.019$} & $0.016$ & $0.071$ & $0.057$ & $0.167$ \\
 \midrule
\multicolumn{11}{l}{$\alpha=1.0$} \\
\midrule
\multirow{5}{*}{RMSE} & VIP & \second{$4.274$} & \best{$8.555$} & \best{$3.032$} & \second{$0.188$} & \second{$0.008$} & \best{$4.075$} & \best{$5.090$} & \best{$0.620$} & \best{$8.770$} \\
 & FTIP & \best{$4.248$} & \second{$8.670$} & \second{$3.100$} & \best{$0.188$} & \best{$0.007$} & \second{$4.100$} & \second{$5.453$} & $0.736$ & \second{$8.860$} \\
 & FBNN & $9.163$ & $14.203$ & $10.317$ & $0.256$ & $0.015$ & $15.358$ & $6.109$ & $0.764$ & $15.859$ \\
 & MFVI & $8.802$ & $14.038$ & $9.641$ & $0.254$ & $0.015$ & $13.610$ & $6.086$ & $0.746$ & $13.385$ \\
 & TFSVI & $6.276$ & $12.245$ & $4.392$ & $0.234$ & $0.014$ & $4.579$ & $5.956$ & \second{$0.671$} & $12.083$ \\
\cmidrule(lr){1-11}
\multirow{5}{*}{NLL} & VIP & \best{$2.68$} & \best{$3.40$} & \second{$2.05$} & \second{$-0.40$} & \second{$-4.23$} & \second{$2.80$} & $2.98$ & \second{$0.93$} & \second{$2.43$} \\
 & FTIP & \second{$2.86$} & \second{$3.41$} & \best{$1.66$} & \best{$-0.41$} & \best{$-4.53$} & \best{$2.80$} & \best{$2.60$} & \best{$-0.85$} & \best{$2.27$} \\
 & FBNN & $3.60$ & $4.08$ & $3.71$ & $0.06$ & $-2.79$ & $4.16$ & $3.27$ & $1.14$ & $4.30$ \\
 & MFVI & $3.60$ & $4.06$ & $3.51$ & $0.05$ & $-2.80$ & $3.80$ & $3.22$ & $1.13$ & $4.06$ \\
 & TFSVI & $3.18$ & $3.91$ & $2.74$ & $-0.04$ & $-2.91$ & $2.93$ & \second{$2.84$} & $1.02$ & $3.23$ \\
\cmidrule(lr){1-11}
\multirow{5}{*}{CRPS} & VIP & \best{$2.264$} & \second{$4.446$} & \best{$1.382$} & \second{$0.100$} & \second{$0.004$} & \best{$2.251$} & \best{$2.788$} & \best{$0.354$} & \second{$3.233$} \\
 & FTIP & \second{$2.311$} & \best{$4.435$} & \second{$1.384$} & \best{$0.099$} & \best{$0.003$} & \second{$2.279$} & \second{$2.903$} & $0.392$ & \best{$3.101$} \\
 & FBNN & $5.010$ & $8.050$ & $5.798$ & $0.148$ & $0.009$ & $9.006$ & $3.577$ & $0.440$ & $9.090$ \\
 & MFVI & $4.852$ & $7.921$ & $5.443$ & $0.145$ & $0.009$ & $7.489$ & $3.563$ & $0.420$ & $7.476$ \\
 & TFSVI & $3.108$ & $6.823$ & $2.319$ & $0.133$ & $0.008$ & $2.553$ & $3.429$ & \second{$0.377$} & $5.275$ \\
\cmidrule(lr){1-11}
\multirow{5}{*}{CQM} & VIP & \second{$0.056$} & $0.042$ & $0.061$ & $0.028$ & \second{$0.045$} & \second{$0.015$} & $0.042$ & \best{$0.024$} & \second{$0.079$} \\
 & FTIP & $0.068$ & $0.039$ & \best{$0.026$} & \best{$0.008$} & $0.047$ & \best{$0.013$} & \best{$0.009$} & $0.143$ & \best{$0.061$} \\
 & FBNN & $0.070$ & $0.033$ & $0.081$ & $0.033$ & $0.061$ & $0.059$ & $0.096$ & $0.104$ & $0.145$ \\
 & MFVI & $0.070$ & \best{$0.016$} & $0.101$ & $0.016$ & $0.072$ & $0.030$ & $0.113$ & $0.106$ & $0.104$ \\
 & TFSVI & \best{$0.040$} & \second{$0.022$} & \second{$0.041$} & \second{$0.013$} & \best{$0.028$} & $0.017$ & \second{$0.027$} & \second{$0.059$} & $0.111$ \\

\bottomrule
\end{tabular}%
}
\end{table}

\subsection{Classification}\label{app:classification}

\begin{table}[t]
\centering
\caption{Binary classification benchmark on SUSY and HIGGS comparing VIP, FTIP, MFVI, FBNN, and TFSVI. Results are averaged over $5$ different random seeds. \textcolor{teal}{\textbf{Best}} and \textcolor{purple}{\textbf{second best}} per column highlighted.}
\label{tab:binary_5k}
\scalebox{0.8}{
\begin{tabular}{lccccc}
\toprule
Model & Accuracy $\uparrow$ & NLL $\downarrow$ & AUC $\uparrow$ & ECE $\downarrow$ & Brier $\downarrow$ \\
\midrule
\multicolumn{6}{l}{\textit{SUSY \scriptsize{(5M instances, 18 features)}}} \\
VIP   & $0.7575 \pm 0.0047$ & $0.7346 \pm 0.0370$ & $0.8242 \pm 0.0039$ & $0.1354 \pm 0.0084$ & $0.1883 \pm 0.0030$ \\
FTIP  & $\textcolor{purple}{\bm{0.8010 \pm 0.0004}}$ & $\textcolor{purple}{\bm{0.4343 \pm 0.0010}}$ & $\textcolor{purple}{\bm{0.8741 \pm 0.0002}}$ & $\textcolor{purple}{\bm{0.0147 \pm 0.0022}}$ & $\textcolor{purple}{\bm{0.1398 \pm 0.0002}}$ \\
MFVI  & $0.7818 \pm 0.0110$ & $0.4741 \pm 0.0213$ & $0.8530 \pm 0.0090$ & $0.0432 \pm 0.0208$ & $0.1552 \pm 0.0085$ \\
FBNN  & $0.5156 \pm 0.0818$ & $0.6570 \pm 0.0241$ & $0.8315 \pm 0.0081$ & $0.1919 \pm 0.0888$ & $0.2340 \pm 0.0122$ \\
TFSVI & $\textcolor{teal}{\bm{0.8019 \pm 0.0008}}$ & $\textcolor{teal}{\bm{0.4294 \pm 0.0010}}$ & $\textcolor{teal}{\bm{0.8752 \pm 0.0006}}$ & $\textcolor{teal}{\bm{0.0059 \pm 0.0017}}$ & $\textcolor{teal}{\bm{0.1392 \pm 0.0003}}$ \\
\midrule
\multicolumn{6}{l}{\textit{HIGGS \scriptsize{(11M instances, 28 features)}}} \\
VIP   & $0.6446 \pm 0.0126$ & $1.4052 \pm 0.2499$ & $0.6939 \pm 0.0107$ & $0.2608 \pm 0.0276$ & $0.2971 \pm 0.0100$ \\
FTIP  & $\textcolor{purple}{\bm{0.6675 \pm 0.0033}}$ & $\textcolor{purple}{\bm{0.6099 \pm 0.0025}}$ & $\textcolor{purple}{\bm{0.7275 \pm 0.0032}}$ & $0.0197 \pm 0.0058$ & $\textcolor{purple}{\bm{0.2108 \pm 0.0012}}$ \\
MFVI  & $0.5298 \pm 0.0000$ & $0.6914 \pm 0.0000$ & $0.5005 \pm 0.0036$ & $\textcolor{teal}{\bm{0.0012 \pm 0.0009}}$ & $0.2491 \pm 0.0000$ \\
FBNN  & $0.4950 \pm 0.0483$ & $0.6921 \pm 0.0057$ & $0.6017 \pm 0.0438$ & $0.0699 \pm 0.0103$ & $0.2495 \pm 0.0028$ \\
TFSVI & $\textcolor{teal}{\bm{0.6968 \pm 0.0035}}$ & $\textcolor{teal}{\bm{0.5779 \pm 0.0041}}$ & $\textcolor{teal}{\bm{0.7661 \pm 0.0045}}$ & $\textcolor{purple}{\bm{0.0116 \pm 0.0052}}$ & $\textcolor{teal}{\bm{0.1970 \pm 0.0017}}$ \\
\bottomrule
\end{tabular}}
\end{table}

\begin{table}[t]
\centering
\caption{Wall-clock time per iteration (ms/iter, 5 seeds) on SUSY and HIGGS.}
\label{tab:binary_iter_times}
\begin{tabular}{lccccc}
\toprule
Dataset & VIP & FTIP & MFVI & FBNN & TFSVI \\
\midrule
SUSY  & $22.9 \pm 0.2$ & $47.5 \pm 0.8$ & $12.2 \pm 0.0$ & $13.0 \pm 0.0$ & $73.9 \pm 7.2$ \\
HIGGS & $23.1 \pm 0.0$ & $46.4 \pm 0.3$ & $12.7 \pm 0.1$ & $14.2 \pm 0.7$ & $92.0 \pm 0.1$ \\
\bottomrule
\end{tabular}
\end{table}

\begin{table}[t]
    \centering
    \small
    \caption{%
    Multiclass classification results on FashionMNIST and CIFAR10. Results are
    reported as mean $\pm$ standard deviation over four seeds. Best results are
    highlighted for each dataset and metric.
    }
    \label{tab:classification}
    \begin{tabular}{llcccc}
        \toprule
        Dataset & Method & Error $\downarrow$ & NLL $\downarrow$ & ECE $\downarrow$ & Brier $\downarrow$ \\
        \midrule
        \multirow{2}{*}{FashionMNIST}
         & VIP  & $\mathbf{0.093} \pm .003$ & $0.503 \pm .036$          & $0.065 \pm .004$          & $0.155 \pm .005$ \\
         & FTIP & $0.094 \pm .003$          & $\mathbf{0.364} \pm .025$ & $\mathbf{0.050} \pm .004$ & $\mathbf{0.147} \pm .004$ \\
        \midrule
        \multirow{2}{*}{CIFAR10}
         & VIP  & $0.354 \pm .008$          & $1.512 \pm .081$          & $0.199 \pm .012$          & $0.541 \pm .013$ \\
         & FTIP & $\mathbf{0.335} \pm .004$ & $\mathbf{1.025} \pm .011$ & $\mathbf{0.079} \pm .006$ & $\mathbf{0.464} \pm .004$ \\
        \bottomrule
    \end{tabular}
\end{table}

We also evaluate \textsc{FTIP} on binary and multiclass classification
benchmarks. These experiments serve a different purpose from the regression
diagnostics in the main text. In classification, the observation likelihood
already introduces a non-Gaussian predictive distribution: Bernoulli likelihoods
encode asymmetric uncertainty over binary labels, and categorical likelihoods
encode multimodal uncertainty over classes. Therefore, the additional
expressiveness introduced by the flow posterior is not primarily designed to
make the predictive distribution more expressive in the same sense as in
regression, where skewness, heavy tails, and multimodality must be represented
directly in the predictive density over continuous targets. Instead, these
experiments test whether the extra flexibility of the flow-transformed
coefficient posterior can improve the behaviour of the \textsc{VIP} surrogate,
especially in terms of calibration and overconfidence.

Table~\ref{tab:binary_5k} reports results on the large-scale binary
classification datasets SUSY and HIGGS~\citep{Baldi2014higgs}. On both datasets, \textsc{FTIP}
substantially improves over \textsc{VIP}. On SUSY, \textsc{FTIP} increases the
accuracy from $0.7575$ to $0.8010$, reduces the NLL from $0.7346$ to $0.4343$,
and improves all remaining metrics, including AUC, ECE, and Brier score. On
HIGGS, the same pattern is observed: \textsc{FTIP} improves the accuracy from
$0.6446$ to $0.6675$, reduces the NLL from $1.4052$ to $0.6099$, and improves
AUC and Brier score. These reductions in NLL are particularly important because
they indicate that the flow posterior partially corrects the severe
overconfidence exhibited by \textsc{VIP}, while at the same time improving
classification accuracy.

Across SUSY and HIGGS, \textsc{FTIP} is consistently the second-best method
behind \textsc{TFSVI} on the main predictive metrics. This comparison should be
interpreted together with the computational structure of the methods.
\textsc{TFSVI} relies on neural-network Jacobians to construct its
function-space approximation, which can become a bottleneck for larger networks
or architectures where Jacobians are expensive or inconvenient to compute.
By contrast, \textsc{FTIP} preserves the sample-forward structure of
\textsc{VIP}: it only requires sampled prior functions and flow-transformed
coefficient samples. Table~\ref{tab:binary_iter_times} reflects this trade-off.
Although \textsc{FTIP} is more expensive per iteration than \textsc{VIP}, it
remains substantially cheaper than \textsc{TFSVI} on both SUSY and HIGGS.

We further evaluate \textsc{VIP} and \textsc{FTIP} on image classification
benchmarks in Table~\ref{tab:classification}. These experiments are included to
test whether \textsc{FTIP} can mitigate the strong overconfidence of
\textsc{VIP} in multiclass classification, as measured by NLL. On
FashionMNIST \citep{Xiao2017fashion}, \textsc{FTIP} obtains a similar error rate to \textsc{VIP}, but
reduces the NLL from $0.503$ to $0.364$, while also improving ECE and Brier
score. On CIFAR10 \citep{Krizhevsky2009cifar}, the improvement is stronger: \textsc{FTIP} reduces the error
from $0.354$ to $0.335$, the NLL from $1.512$ to $1.025$, the ECE from $0.199$
to $0.079$, and the Brier score from $0.541$ to $0.464$. Thus, even though the
flow posterior is not designed to add output-space multimodality to
classification likelihoods, it still improves the quality of the posterior
approximation induced by the \textsc{VIP} surrogate.

Overall, the classification results show that the benefit of \textsc{FTIP} in
classification is mainly calibration-oriented rather than a direct increase in
the expressive form of the categorical predictive likelihood. The likelihoods
used in binary and multiclass classification already encode non-Gaussian,
asymmetric, and multimodal predictive distributions over labels. Nevertheless,
the flow-transformed coefficient posterior improves the variational
approximation relative to the Gaussian posterior used by \textsc{VIP}. This
leads to better NLL and calibration, and in several cases also better accuracy,
while retaining a more scalable sample-forward inference mechanism than
Jacobian-based function-space methods such as \textsc{TFSVI}.

\subsection{Relation between Gaussian CRPS and RMSE}
\label{app:crps}

For a Gaussian predictive distribution
\begin{equation}
    p(y\mid x)=\mathcal N\big(\mu(x),\sigma^2(x)\big),
\end{equation}
the continuous ranked probability score (\textsc{CRPS}) for a single observation $y$ has the closed form
\begin{equation}
    \mathrm{CRPS}
    \big(
        \mathcal N(\mu,\sigma^2),y
    \big)
    =
    \sigma
    \left[
        z\big(2\Phi(z)-1\big)
        +
        2\phi(z)
        -
        \frac{1}{\sqrt{\pi}}
    \right],
    \qquad
    z=\frac{y-\mu}{\sigma},
    \label{eq:gaussian-crps}
\end{equation}
where $\phi$ and $\Phi$ denote the standard normal density and distribution function, respectively. Thus, \textsc{CRPS} depends not only on the prediction error $y-\mu$, but also on the predictive standard deviation $\sigma$. Consequently, it is not generally a fixed multiple of the root mean squared error (\textsc{RMSE}).

A simple proportional relationship arises only in the calibrated homoscedastic Gaussian case. Suppose that the predictive distribution is correctly specified, so that
\begin{equation}
    y-\mu \sim \mathcal N(0,\sigma^2).
\end{equation}
Then the expected Gaussian \textsc{CRPS} satisfies
\begin{equation}
    \mathbb E
    \left[
        \mathrm{CRPS}
        \big(
            \mathcal N(\mu,\sigma^2),Y
        \big)
    \right]
    =
    \frac{\sigma}{\sqrt{\pi}}.
    \label{eq:expected-gaussian-crps}
\end{equation}
In the same setting,
\begin{equation}
    \mathrm{RMSE}
    =
    \sqrt{\mathbb E[(Y-\mu)^2]}
    =
    \sigma.
\end{equation}
Therefore,
\begin{equation}
    \mathbb E[\mathrm{CRPS}]
    =
    \frac{1}{\sqrt{\pi}}
    \mathrm{RMSE}
    \approx
    0.564\,\mathrm{RMSE}.
    \label{eq:crps-rmse-ratio}
\end{equation}

This relationship is only a useful calibration heuristic. When the predictive variance is misspecified, heteroscedastic, biased, or non-Gaussian, the ratio between \textsc{CRPS} and \textsc{RMSE} need not be close to $1/\sqrt{\pi}$. In particular, \textsc{CRPS} remains sensitive to the full predictive distribution, whereas \textsc{RMSE} depends only on the predictive mean.


\end{document}